\documentclass[lettersize,journal]{IEEEtran}
\usepackage{amsmath,amsfonts}
\usepackage{algorithmic}
\usepackage{algorithm}
\usepackage{array}
\usepackage[caption=false,font=normalsize,labelfont=sf,textfont=sf]{subfig}
\usepackage{textcomp}
\usepackage{stfloats}
\usepackage{url}
\usepackage{verbatim}
\usepackage{graphicx}
\usepackage{cite}
\usepackage{booktabs} 
\usepackage{xcolor}
\usepackage{adjustbox}
\usepackage{multirow}
\usepackage{bbding}
\usepackage{pifont}
\usepackage{utfsym}
\usepackage{colortbl}
\usepackage{makecell}

\newcommand{\zs}[1]{\textcolor{black}{#1}}

\newcommand{\re}[1]{\textcolor{black}{#1}}

\makeatletter
\def\IEEEkeywordsname{Keywords}
\makeatother

\begin{document}

\title{DPMambaIR: All-in-One Image Restoration via Degradation-Aware Prompt State Space Model}

\author{{Zhanwen~Liu,~Sai~Zhou,~Yuchao~Dai, ~Yang~Wang$^{\dagger}$,~Yisheng~An,~Xiangmo~Zhao}
\thanks{$^{\dagger}$Corresponding author.}
\thanks{Zhanwen Liu, Sai Zhou, Yang Wang, Yisheng An and Xiangmo Zhao are with the School of Information Engineering, Chang'an University, Shaanxi, Xi’an 710000, China (e-mail: zwliu@chd.edu.cn; 2024124095@chd.edu.cn; ywang120@chd.edu.cn, aysm@chd.edu.cn, xmzhao@chd.edu.cn).}
\thanks{Yuchao Dai is with the School of Electronics and Information, Northwestern
 Polytechnical University, Xi’an 710072, China (e-mail: daiyuchao@gmail.com).}
}

\maketitle

\begin{abstract}

All-in-One image restoration aims to address multiple image degradation problems using a single model, offering a more practical and versatile solution compared to designing dedicated models for each degradation type. Existing approaches typically rely on Degradation-specific models or coarse-grained degradation prompts to guide image restoration. However, they lack fine-grained modeling of degradation information and face limitations in balancing multi-task conflicts. To overcome these limitations, we propose DPMambaIR, a novel All-in-One image restoration framework that introduces a fine-grained degradation extractor and a Degradation-Aware Prompt State Space Model (DP-SSM). The DP-SSM leverages the fine-grained degradation features captured by the extractor as dynamic prompts, which are then incorporated into the state space modeling process. This enhances the model's adaptability to diverse degradation types, while a complementary High-Frequency Enhancement Block (HEB) recovers local high-frequency details. Extensive experiments on a mixed dataset containing seven degradation types show that DPMambaIR achieves the best performance, with 27.69 dB and 0.893 in PSNR and SSIM, respectively. These results highlight the potential and superiority of DPMambaIR as a unified solution for All-in-One image restoration.


\end{abstract}

\section{Introduction}
\IEEEPARstart{I}{mage} restoration is a fundamental task in computer vision, aiming to recover high-quality images from degraded inputs. Recent advances in deep learning have significantly improved performance in degradation-specific restoration tasks, such as image denoising \cite{tmmdenoising1, kim2024lan}, deblurring \cite{tmmdeblur1}, deraining\cite{tmmderain1}, dehazing\cite{tmmdehaze, wang2024odcr}, and low-light enhancement\cite{tmmlol}. However, traditional methods typically rely on degradation-specific models explicitly tailored to individual degradation types. These approaches often require prior knowledge of the degradation type, which limits their practicality in real-world scenarios, such as autonomous driving and nighttime surveillance, where diverse and unknown degradations frequently coexist.

\begin{figure}
    \centering
    \includegraphics[width=0.95\linewidth]{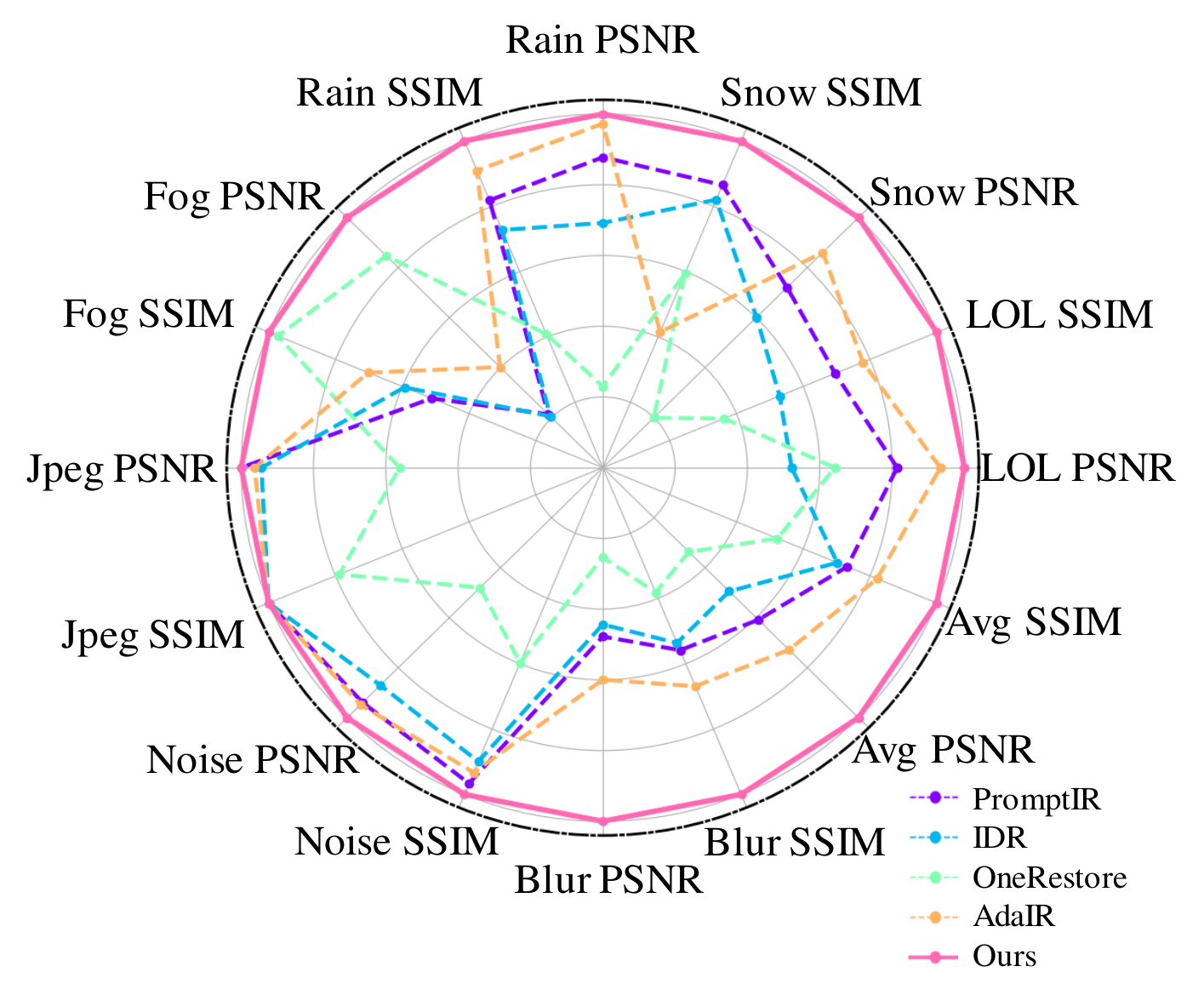}
    \vspace{-10pt} 
    \caption{Comparison with SOTA All-in-One Image Restoration Methods on mixed Dataset. Our DPMambaIR achieves consistently better performance.}
    \label{fig:fig1}
    \vspace{-10pt}
\end{figure}

To address this limitation, recent studies explore All-in-One image restoration frameworks that unify multiple restoration tasks into a single model. Existing methods can be broadly categorized into Mixture-of-Experts (MoE)-based and Prompt-based approaches. MoE-based methods, such as MoFME~\cite{zhang2024MoFME}, dynamically route tasks to specialized experts using an uncertainty-aware router. However, these methods suffer from high computational costs and lack explicit degradation modeling, limiting adaptability to unseen degradations. Alternatively, Prompt-based methods like OneRestore~\cite{guo2024onerestore} guide restoration using degradation prompts, integrating scene descriptors via cross-attention mechanisms.

\re{However, existing prompt-based strategies~\cite{guo2024onerestore, promptir} suffer from fundamental limitations in how degradation information is represented and exploited. A key shortcoming lies in their reliance on static priors, where degradations are encoded as discrete class labels or fixed semantic tokens. Such categorical representations are inherently inadequate for modeling the continuous nature of real-world degradations and their spatial variability, for example, differentiating between light haze and dense, spatially accumulating fog. Furthermore, most of these methods follow an additive fusion paradigm in which degradation cues are treated merely as auxiliary information appended to visual features. This mechanism is intrinsically limited, as the underlying network architecture remains governed by fixed convolutional kernels or static attention patterns. Consequently, the model is compelled to address heterogeneous restoration demands, such as suppressing high-frequency noise while simultaneously enhancing global contrast, within a rigid geometric structure. This conflict ultimately degrades performance across diverse and complex degradation scenarios.
}

\re{To address these challenges, we propose \textbf{DPMambaIR}, a novel All-in-One framework that shifts the design philosophy from conventional feature fusion to explicit \textit{dynamic parameter modulation}. Our approach is built upon a fine-grained degradation extractor and a Degradation-Aware Prompt State Space Model (DP-SSM). Unlike methods that rely on discrete degradation labels, the proposed extractor employs a reconstruction objective to regress continuous degradation embeddings, enabling it to encode nuanced information regarding both degradation type and severity. These embeddings are utilized to directly modulate the core State Space Model parameters $(\Delta, B, C)$. Modulating the discretization step $\Delta$ effectively alters the integration step size of the underlying ordinary differential equation, allowing the model to realize distinct continuous dynamics conditioned on degradation characteristics. Specifically, the network learns to assign smaller $\Delta$ values to induce a high-inertia state favorable for smoothing high-frequency noise, whereas larger values are generated to enter a high-gain regime that amplifies weak signals in low-light conditions. Concurrently, the input projection matrix $B$ is modulated to control how strongly external observations drive the latent state transitions, while the output matrix $C$ is adjusted to inject global degradation priors. Furthermore, we observe that jointly optimizing heterogeneous degradations often induces an implicit bias toward low-frequency structures, making the consistent recovery of fine details challenging. To alleviate this issue, we incorporate a lightweight High-frequency Enhancement Block (HEB) that complements the proposed dynamic modulation and facilitates the restoration of local textures.}

Extensive experiments on a mixed dataset containing seven degradation types demonstrate the efficacy of DPMambaIR. As shown in Fig.~\ref{fig:fig1}, DPMambaIR achieves state-of-the-art performance in terms of PSNR and SSIM, outperforming existing All-in-One methods such as AdaIR and OneRestore. Moreover, DPMambaIR achieves competitive results across individual tasks, including deraining, low-light enhancement, deblurring, and dehazing, demonstrating its potential as an effective and robust All-in-One image restoration solution.

The contributions of this paper are summarized as follows:
\begin{itemize}
\item[$\bullet$] \re{We propose a degradation extractor capable of capturing fine-grained, continuous degradation features from complex degraded images via a regression-based reconstruction objective.}
\item[$\bullet$] \re{We design a Degradation-aware Prompt State Space Model (DP-SSM) that introduces a dynamic parameter modulation mechanism. By dynamically modulating the state-space evolution parameters based on degradation priors, it enhances the model's physical adaptability to diverse degradation dynamics.}
\item[$\bullet$] Extensive experiments on a mixed dataset containing seven types of degradation demonstrate that DPMambaIR achieves state-of-the-art performance on PSNR and SSIM metrics, outperforming existing methods. The results validate the effectiveness and robustness of the proposed approach.
\end{itemize}
\section{Related Work}

\subsection{All-in-One Image Restoration}
Image restoration remains a foundational challenge in computer vision, striving to reconstruct high-fidelity content from corrupted inputs. While traditional approaches, such as Dark Channel Prior~\cite{he2010single} and Color Line Prior~\cite{fattal2014dehazing}, relied on handcrafted heuristics that often struggle in complex scenarios, the advent of deep learning~\cite{lecun2015deep} has revolutionized the field. Significant strides have been made in specific domains, including super-resolution~\cite{tmmsr1, peng2024towards, peng2025pixel}, denoising~\cite{tmmdenoising1, kim2024lan}, deblurring~\cite{tmmdeblur1, zhang2018adversarial, zhang2023mc}, deraining~\cite{tmmderain1, zhang2022enhanced}, dehazing~\cite{tmmdehaze, jin2025mb}, desnowing~\cite{guo2025deep, zhang2021deep}, and low-light enhancement~\cite{tmmlol}. Leveraging CNN-based~\cite{lecun1989handwritten}, Transformer-based~\cite{vaswani2017attention}, or diffusion-based~\cite{ho2020denoising} architectures, these methods excel at feature representation. However, their reliance on degradation-specific designs fundamentally limits their utility in real-world environments where degradations are diverse, compounded, and often unknown.


Recent research has consequently pivoted towards All-in-One frameworks that emphasize flexibility and unified processing. These methods generally employ multi-task learning strategies, such as Mixture-of-Experts (MoE) or prompt-guided adaptation. For instance, AirNet~\cite{airnet} utilizes contrastive learning to derive degradation representations, while MoFME~\cite{zhang2024MoFME} implements an uncertainty-aware router to dynamically dispatch features to specialized experts. Although effective, MoE architectures often incur high computational overhead and lack explicit physical modeling of the degradation. In parallel, prompt-based methods such as PromptIR~\cite{promptir}, OneRestore~\cite{guo2024onerestore}, and DA-CLIP~\cite{daclip} integrate degradation cues, ranging from learned visual prompts to text embeddings, via cross-attention mechanisms. IDR~\cite{idr} further explores meta-learning for decomposition, and AdaIR~\cite{cui2025adair} adopts frequency domain decoupling for adaptive restoration.

\re{Despite these advancements, fundamental limitations persist in the representation and exploitation of degradation information. Most prompt-based strategies~\cite{promptir, guo2024onerestore} rely on coarse-grained priors or discrete class labels, which fail to capture the continuous nature and spatial variability inherent in real-world images. Furthermore, they typically adopt an additive fusion paradigm where degradation cues are treated merely as auxiliary inputs. This mechanism is intrinsically limited because the underlying network parameters remain fixed, governed by static weights or attention patterns that lack the flexibility to adapt to heterogeneous restoration demands. In contrast, our DPMambaIR framework employs a regression-based extractor to derive fine-grained, continuous degradation embeddings. We depart from simple feature concatenation and utilize these embeddings to dynamically re-parameterize the State Space Model. This shift from additive fusion to explicit dynamic parameter modulation allows the system to adapt its evolution dynamics to the specific physical properties of the degradation.}

\begin{figure*}
    \centering
    \includegraphics[width=1\linewidth]{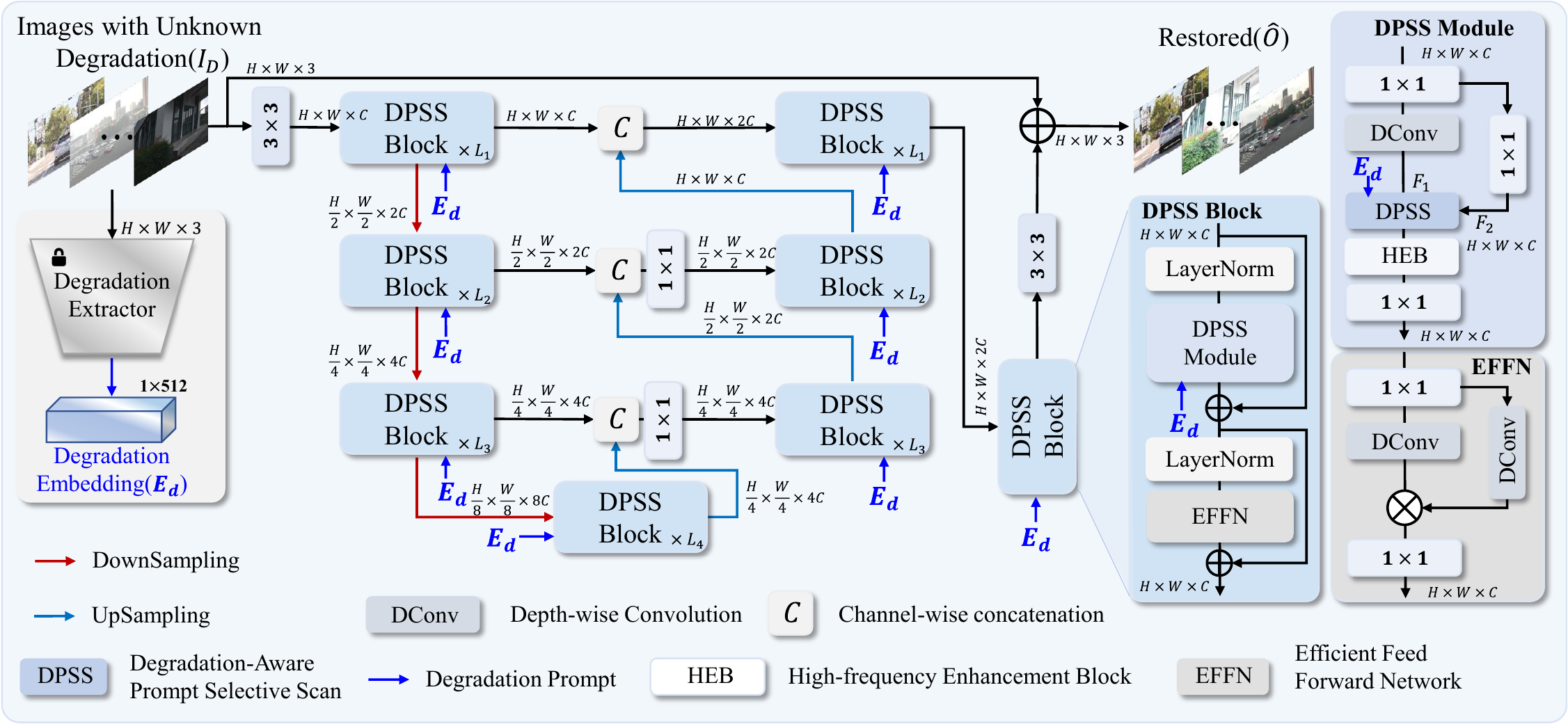}
    \caption{
      The overall framework of DPMambaIR consists of a shallow feature extractor and multiple DPSS blocks, with an upsampler implemented by pixel shuffle and a downsampler by pixel unshuffle. Additionally, a degradation extractor is included to extract degradation embeddings, which are then inserted into the DPSS blocks.
    }
    \label{fig:fig2}
\end{figure*}

\subsection{State Space Model based Image Restoration}

Convolutional Neural Networks (CNNs)~\cite{lecun2015deep} and Transformers~\cite{vaswani2017attention} have long dominated image restoration. While CNNs excel at capturing local high-frequency details, their limited receptive fields hinder the modeling of long-range dependencies essential for correcting global degradations like haze or uneven lighting. Transformers address this with global self-attention but suffer from quadratic computational complexity relative to image resolution.

To reconcile long-range modeling with efficiency, researchers have revisited the State Space Model (SSM)~\cite{kalman1960new}. SSMs map a 1D input sequence $x(t)$ to an output $y(t)$ via a latent state $h(t) \in \mathbb{R}^N$, governed by the linear ordinary differential equation (ODE):
\begin{equation}
    \begin{aligned}
        h'(t) &= Ah(t) + Bx(t), \\
        y(t) &= Ch(t) + Dx(t).
    \end{aligned}
    \label{eq:1}
\end{equation}
where $A, B, C, D$ are learnable parameters. For digital implementation, this continuous system is discretized with a step size $\Delta$, yielding the recurrence:
\begin{equation}
    \begin{aligned}
        h_i &= \overline{A} h_{i-1} + \overline{B} x_i, \\
        y_i &= Ch_i + Dx_i.
    \end{aligned}
    \label{eq:2}
\end{equation}
Here, the discretized matrices are defined as $\overline{A} = \exp(\Delta A)$ and $\overline{B} = (\Delta A)^{-1} (\exp(\Delta A) - I)\Delta B$. This formulation allows for linear-complexity inference while maintaining a global receptive field.

The Vision Mamba~\cite{vim} first adapted SSMs for computer vision, demonstrating that the Vision State Space Module could outperform Vision Transformers~\cite{vit} with significantly lower computational costs. This success has sparked adoption across tasks including object detection~\cite{yang2025smamba}, segmentation~\cite{xiao2024spatial}, and classification~\cite{sheng2024dualmamba}. In image restoration, MambaIR~\cite{guo2024mambair} pioneered the use of SSMs to achieve a balance between efficiency and perceptual quality, inspiring subsequent variants~\cite{guo2024mambairv2, peng2025directing, di2024qmambabsr}.

However, applying standard SSMs to All-in-One restoration faces two primary limitations. First, \textbf{Limited Global Context Utilization}: The causal nature of standard scanning means the $i$-th pixel only accesses information from preceding pixels. While multi-directional scanning can alleviate this, it increases computation without always yielding proportional gains in low-level tasks~\cite{guo2024mambair}. Second, and more critically, \textbf{Lack of Degradation-Awareness}: Existing SSM-based restoration methods rely on fixed transition dynamics ($A, B, C$) or purely input-driven modulation. They lack an explicit mechanism to adapt the system's evolution rules based on external degradation priors (e.g., adjusting the integration step $\Delta$ for different noise levels). Consequently, their ability to generalize across the diverse degradation types encountered in All-in-One settings remains constrained.

Motivated by these gaps, we propose the Degradation-aware Prompt State Space Model (DP-SSM). Unlike prior works, our method utilizes fine-grained degradation embeddings to dynamically modulate the discretization step $\Delta$, along with the state-space transition and output matrices. This design establishes a dynamics-aware mechanism that harmonizes global context modeling with degradation-specific physical guidance, significantly enhancing robustness in complex restoration scenarios.
\section{Method}
\subsection{Problem Definition}
All-in-One Image Restoration aims to address a wide range of image degradation types within a unified framework. Specifically, given a degraded image $\boldsymbol{I}_{D} \in \mathbb{R}^{3 \times H \times W}$, where $D$ denotes the degradation process, our goal is to restore the clean image $\boldsymbol{\hat{O}} \in \mathbb{R}^{3 \times H \times W}$ using a unified model $\mathcal{F}$. This can be formulated as:
\begin{equation}
    \begin{aligned}
        \boldsymbol{\hat{O}} = \mathcal{F}(\boldsymbol{I}_{D}).
    \end{aligned}
    \label{eq:3}
\end{equation}
where $\mathcal{F}$ is designed to generalize across various degradation types, such as noise, blur, compression artifacts, and weather effects, without explicit knowledge of $Deg$. The degradation process $Deg$ can be mathematically represented as:
\begin{equation}
    \begin{aligned}
        \boldsymbol{I}_{D} = Deg(\boldsymbol{O}) + \boldsymbol{\eta}.
    \end{aligned}
    \label{eq:4}
\end{equation}
Here, $\boldsymbol{O} \in \mathbb{R}^{3 \times H \times W}$ denotes the original clean image, $Deg$ represents the degradation operator (\textit{e.g.}, blur kernel, brightness, or quality compression), and $\boldsymbol{\eta}$ signifies additive noise. The main challenge arises from the fact that $Deg$ is usually unknown and varies across tasks, making it difficult to design a universal model capable of effectively handling all possible degradations.

Unlike degradation-specific restoration, the All-in-One setting requires the model to generalize across a broad spectrum of degradation types without prior knowledge of degradation parameters. This introduces two main challenges: (a) how to represent fine-grained degradation information in a unified way; and (b) how to dynamically adapt the model behavior based on such information during inference.

\begin{figure*}
    \centering
    \includegraphics[width=1\linewidth]{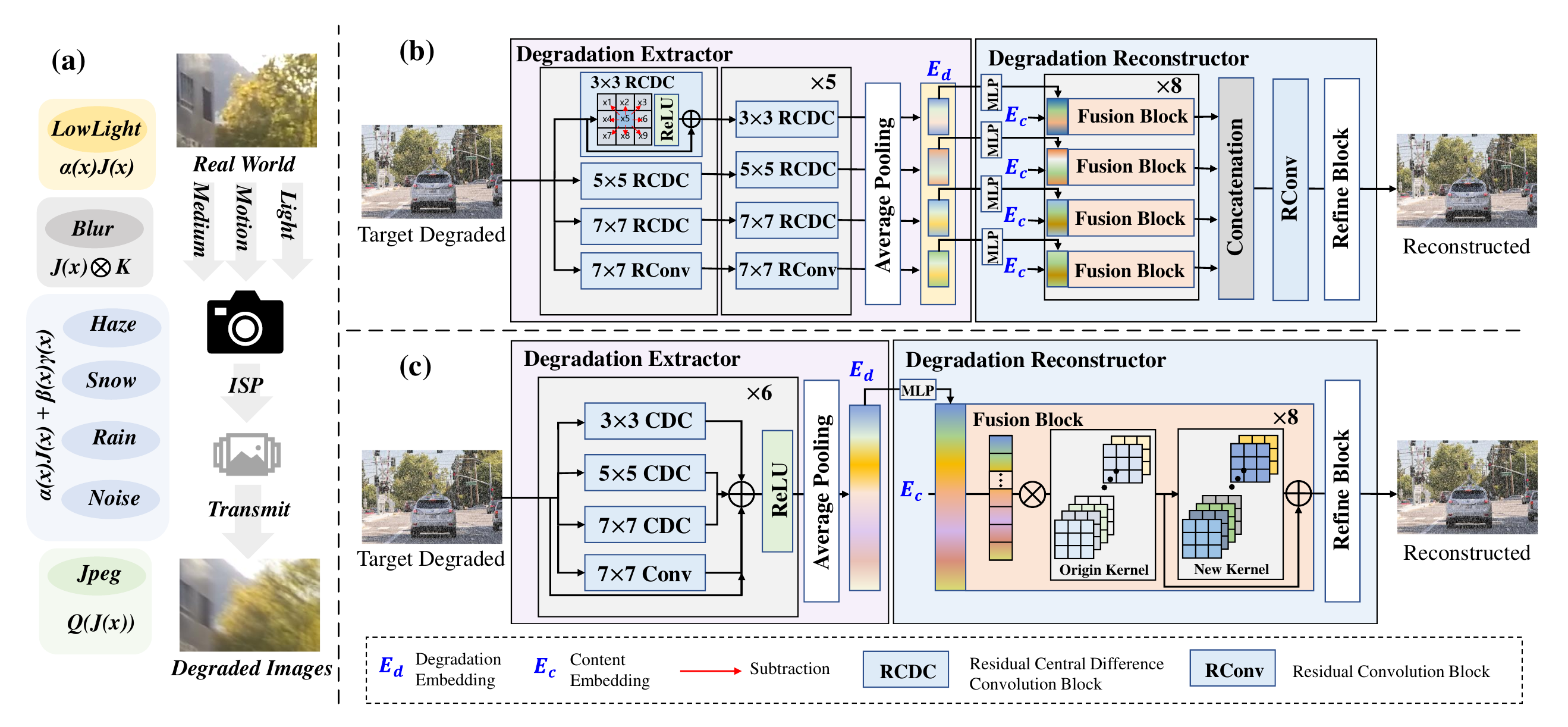}
    \caption{(a) Real-World Image Degradation Modeling Process. (b) Architecture details of the Large version degradation extractor and reconstructor. (c) Architecture details of the normal version degradation extractor and reconstructor.}
    \label{fig:fig9}
\end{figure*}

\subsection{DPMambaIR}


We propose \textbf{DPMambaIR}, a degradation-aware framework for All-in-One image restoration. As illustrated in Fig.~\ref{fig:fig2}, our method is built upon a Mamba-based U-shaped architecture with an asymmetric encoder-decoder design. Initially, a $3 \times 3$ convolution extracts shallow features $\boldsymbol{F} \in \mathbb{R}^{H \times W \times C}$ from the degraded input. The encoder comprises three stages, each utilizing downsampling to progressively halve the feature resolution while doubling the channel dimension. Correspondingly, the decoder employs upsampling and skip connections to fuse multi-scale features from the encoding path. Finally, the clean image is reconstructed via a refinement stage, utilizing a global residual connection to superimpose the learned residual onto the degraded input.


To address the challenges posed by unknown degradation types and severities in All-in-One image restoration, we propose a \textit{Degradation Extractor} pre-trained via a self-supervised reconstruction objective. This module encodes heterogeneous degradation patterns into a compact embedding, effectively bridging the gap between blind and non-blind restoration by providing explicit degradation representations. To fully exploit these representations, we revisit the selective state-space modeling paradigm. Specifically, we propose a \textit{Degradation-Aware Prompt State Space Model}, which dynamically modulates the restoration trajectory by injecting global degradation priors into the state transition process.

\subsection{Degradation Extractor}
To effectively extract degradation priors, we formulate a generalized degradation model that encapsulates the physical corruption process from image capture to transmission, as shown in Fig.~\ref{fig:fig9} (a). This process typically involves four degradation categories: (1) illumination degradation, (2) motion-induced blur, (3) transmission medium artifacts, and (4) compression artifacts. We unify these factors into a composite formation model:
\begin{equation}
    \begin{aligned}
        I(x) = Q\big((\alpha(x)J(x) + \beta(x)\gamma(x)) \otimes K\big).
    \end{aligned}
    \label{eq:5}
\end{equation}
where $J(x)$ denotes the latent clean image. The term $\alpha(x)$ models luminance variations, encompassing low-light conditions or uneven illumination caused by weather. $\gamma(x)$ represents additive artifacts introduced by the transmission medium, such as rain streaks, snow, or haze aerosols, while $\beta(x)$ modulates the spatial intensity of these artifacts. $K$ denotes the blur kernel resulting from relative motion, and $Q$ represents quantization or compression operations applied during storage.


Guided by this formulation, we design a degradation extractor capable of capturing heterogeneous corruption patterns, as illustrated in Fig.~\ref{fig:fig9}(b). Specifically, we employ Central Difference Convolution (CDC)~\cite{wang2023decoupling, peng2024efficient, peng2024lightweight} to identify gradient-based degradation cues, such as edges and textures associated with blur and noise. A multi-scale module aggregates features from four parallel CDC branches with varying kernel sizes, enabling the perception of degradations across different frequency bands. Additionally, we integrate a standard convolution branch to capture global intensity and brightness distributions.

To balance extraction accuracy with computational efficiency, we employ a re-parameterization technique during inference, merging the multi-branch structure into a compact representation, as shown in Fig.~\ref{fig:fig9} (c). This yields two model variants: a larger capacity DPMambaIR-L utilizing the full multi-branch structure, and a lightweight DPMambaIR employing the re-parameterized extractor (detailed in Table~\ref{tab:tab2}).

To enable the learning of fine-grained degradation embeddings, we propose a reconstruction-based pre-training strategy. Unlike classification-based methods that rely on discrete labels, this approach forces the extractor to learn a continuous manifold of degradation types and severities. Formally, the extractor $\mathcal{E}$ maps a degraded input $\boldsymbol{I}_{D}$ to a low-dimensional embedding $\boldsymbol{E}_{d}$:
\begin{equation}
    \boldsymbol{E}_{d} = \mathcal{E}(\boldsymbol{I}_{D}).
\end{equation}

To supervise this process, a reconstruction decoder $\mathcal{R}$ attempts to recover the degraded image $\boldsymbol{I}_{D}$ by combining $\boldsymbol{E}_{d}$ with a content embedding $\boldsymbol{E}_{c}$ derived from the corresponding clean ground truth $\boldsymbol{O}$:

\begin{equation}
    \boldsymbol{E}_{c} = \text{Restormer}(\boldsymbol{O}), \ \ 
    \boldsymbol{\hat{I}}_{D} = \mathcal{R}(\boldsymbol{E}_{c}, \boldsymbol{E}_{d}).
\end{equation}
\re{This disentanglement ensures that $\boldsymbol{E}_{d}$ captures pure degradation information necessary to reconstruct the corruption, independent of the image content.} The optimization minimizes a hybrid objective combining $L_1$ loss and the Learned Perceptual Image Patch Similarity (LPIPS) metric, defined as $\mathcal{L} = \|\boldsymbol{I}_{D} - \boldsymbol{\hat{I}}_{D}\|_1 + \lambda \mathcal{L}_{\text{LPIPS}}$, to ensure perceptually faithful reconstruction of the degradation patterns.


\begin{figure*}
    \centering
    \includegraphics[width=0.9\linewidth]{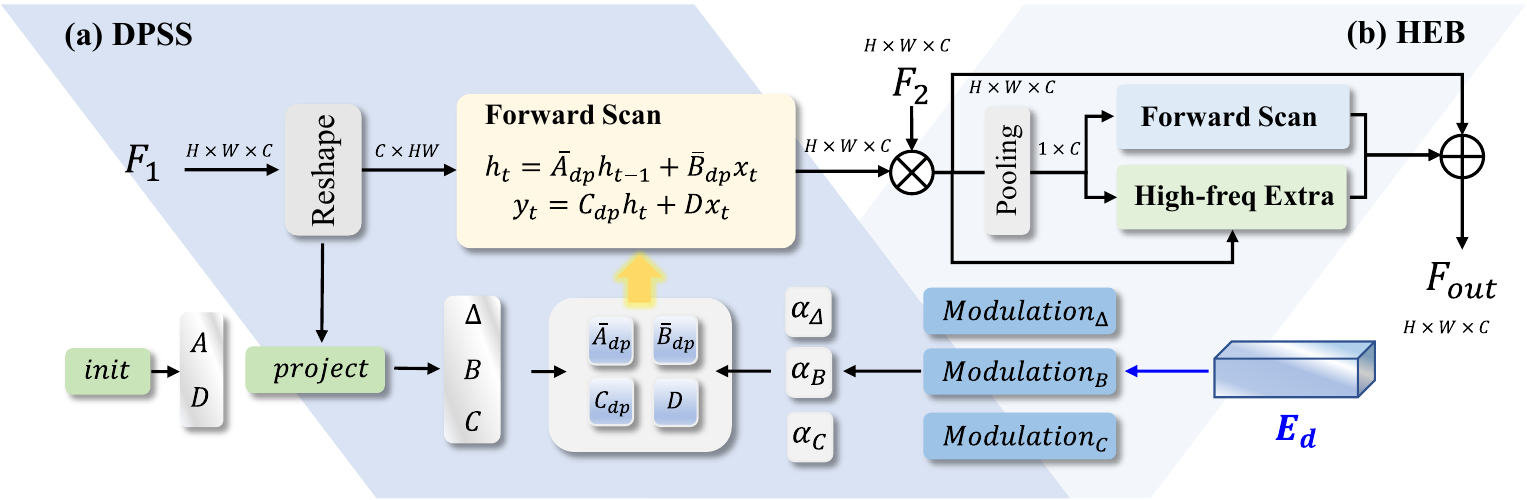}
    \caption{The Architecture of (a) Degradation-aware Prompt Selective Scan (DPSS) and (b) High-frequency Enhancement Block (HEB), with Three Inputs \(F_1\), \(F_2\), and Degradation Embedding \(\boldsymbol{E}_{d}\).}
    \label{fig:fig4}
\end{figure*}

\subsection{Degradation-aware Prompt State Space Model}
\zs{To address the limitations of SSM-based image restoration, specifically the insufficient use of global context and the lack of effective degradation-aware mechanisms,} we propose a novel Degradation-aware Prompt State Space Model (DP-SSM) to effectively adapt SSM to All-in-One image restoration. Specifically, we propose a novel Degradation-aware Prompt Selective Scan, which incorporates degradation information into the State Space Model. This is achieved by leveraging a Degradation Embedding extracted from a pre-trained Degradation Extractor to provide supplementary degradation-related contextual information that the traditional SSM cannot capture. In addition, we modulate the parameter matrices in the state-space equations to establish a degradation-aware mechanism. 

This mechanism enables SSM to adapt to \zs{multi-task learning and handle task competition.}
\zs{To better recover local high-frequency details}, we design a lightweight High-frequency Enhancement Block (HEB). This component ensures the preservation of high-frequency details and enhances the overall model performance.

\noindent\textbf{Degradation-aware Prompt Selective Scan. }
We revisit the traditional SSM, whose parameter settings are shown in Eq.(\ref{eq:2}). Matrices \(A\), \(B\), \(C\), and \(D\) control the SSM output. The state transition matrix \(A\) compresses historical information, while the input matrix \(B\) maps input signals to the state space, determining their influence on historical states. The output matrix \(C\) maps historical states to the observable space, reflecting their impact on the output. Matrix \(D\) provides a direct pathway from input to output, similar to a residual connection. In traditional SSM, these matrices and \(\Delta\) are derived from:

\begin{equation}
\begin{aligned}
    &A=init(),D=init(),\\
    &B=Linear(x),C=Linear(x),\Delta = Linear(x).
\end{aligned}
\end{equation}
where $init()$ is an initialization method that does not depend on the input $x$. 
However, this input-dependent design is agnostic to the degradation type. It processes high-frequency noise and low-frequency blur with the same initialization prior, lacking the ability to structurally adapt the system dynamics for specific restoration tasks.
This motivates the introduction of a degradation-aware prompt mechanism, which guides the state-space modeling by incorporating specific degradation cues.

\zs{To integrate degradation awareness into the SSM,} we first utilize a degradation extractor \(\mathcal{E}\) to extract the degradation embedding \(\boldsymbol{E}_{d}\) from the input image \(I_D\). \zs{This embedding is then employed to modulate \(\Delta\), the input matrix \(B\), and the output matrix \(C\), thereby enhancing the model’s ability to compress historical information, process input signals, and capture global context, as illustrated in Fig.~\ref{fig:fig4} (a).} The specific formulation is given as follows:
\begin{equation}
    \begin{aligned}
        \boldsymbol{E}_{d} &= \mathcal{E}(I_D), \\
        \alpha_\Delta &= \mathcal{M}_\Delta(\boldsymbol{E}_{d}), \,
        \alpha_B = \mathcal{M}_B(\boldsymbol{E}_{d}), \,
        \alpha_C = \mathcal{M}_C(\boldsymbol{E}_{d}), \\
        \Delta_{dp} &= \alpha_\Delta \cdot \Delta, \,
        B_{dp} = \alpha_B \cdot B, \,
        C_{dp} = \alpha_C \cdot C.
    \end{aligned}
    \label{eq:9}
\end{equation}
Here, \(\mathcal{M}_\Delta\), \(\mathcal{M}_B\), and \(\mathcal{M}_C\) are combinations of linear layers used to map the degradation embedding \(\boldsymbol{E}_{d}\) into feature spaces, forming modulation vectors \(\alpha_\Delta\), \(\alpha_B\), and \(\alpha_C\) for \(\Delta\), \(B\), and \(C\), respectively. \(\Delta_{dp}\), \(B_{dp}\), and \(C_{dp}\) are the newly modulated SSM parameters.

Finally, the formulation of the Degradation-aware State Space Model can be expressed as follows:
\begin{equation}
    \begin{aligned}
    \overline{A}_{dp} &= \exp(\Delta_{dp} A), \\
    \overline{B}_{dp} &= (\Delta_{dp} A)^{-1} (\exp(\Delta_{dp}A) - I) \cdot \Delta_{dp} B_{dp}, \\ 
    h_i &= \overline{A}_{dp_i} h_{i-1} + \overline{B}_{dp_i} x_i, \\
    y_i &= C_{dp} h_i + D x_i.
    \end{aligned}
\end{equation}

\begin{figure}[t]
    \centering
        \includegraphics[width=\linewidth]{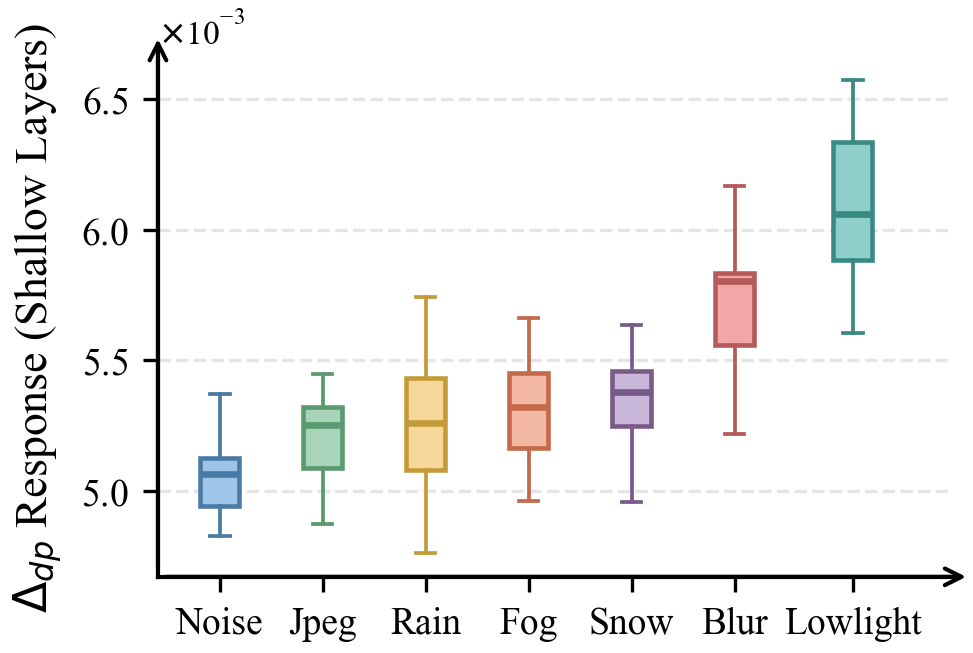} 
    \caption{\re{Distribution of the learned parameter $\Delta_{dp}$ across different degradation types in the first 20\% of DPSS layers. The model adaptively assigns smaller step sizes for noise/jpeg and larger step sizes for blur/low-light, confirming physically interpretable dynamic modulation.}}
    \label{fig:delta_stats}
\end{figure}

\re{The proposed modulation mechanism fundamentally alters the underlying system dynamics, effectively recalibrating the trade-off between memory inertia and input sensitivity. Central to this adaptation is the step size $\Delta_{dp}$, whose learned distribution shown in Fig.~\ref{fig:delta_stats} reveals a distinct physical correspondence to the degradation characteristics. For high-frequency degradations such as noise and JPEG artifacts, the model autonomously converges to a minimal $\Delta_{dp}$ regime. As $\Delta_{dp}$ approaches zero, the discretized transition matrix $\overline{{A}}_{dp}$ approximates the identity matrix while the input gain $\overline{{B}}_{dp}$ diminishes. This configuration transforms the SSM into a stable low-pass filter, creating a high-inertia state that effectively suppresses instantaneous pixel variance by prioritizing historical context over the unreliable current observation.}
\re{In contrast, inputs characterized by low-frequency or structural deficiencies, such as blur and low-light conditions, trigger a marked increase in the learned $\Delta_{dp}$. This shift transitions the system into a high-gain regime, amplifying the magnitude of $\overline{{B}}_{dp}$ to enhance sensitivity towards weak input signals and facilitate the recovery of attenuated local structures.
Complementing this dynamic discretization, the concurrent modulation of ${B}$ acts as a gating mechanism for input observations, while the modulation of ${C}$ supplements the feature readout with global degradation cues.}

\re{Through this strategy, the transformation matrix ${A}$ and input matrix ${B}$ are endowed with degradation-aware capabilities, allowing them to adaptively regulate memory inertia and input sensitivity according to the specific degradation severity. However, the standard SSM remains constrained by its strictly causal nature, where the $i$-th pixel can only access information from the preceding $i-1$ pixels, resulting in a restricted global receptive field. Although existing methods attempt to alleviate this via parallel multi-directional scanning, such approaches incur high computational costs and yield limited performance gains in low-level vision tasks~\cite{guo2024mambair}. By retaining the efficient single-directional scanning and instead modulating matrix ${C}$ to inject global degradation information, our method achieves effective global degradation awareness with significantly lower computational overhead.}

\begin{figure}[t]
    \centering
    \includegraphics[width=0.9\linewidth]{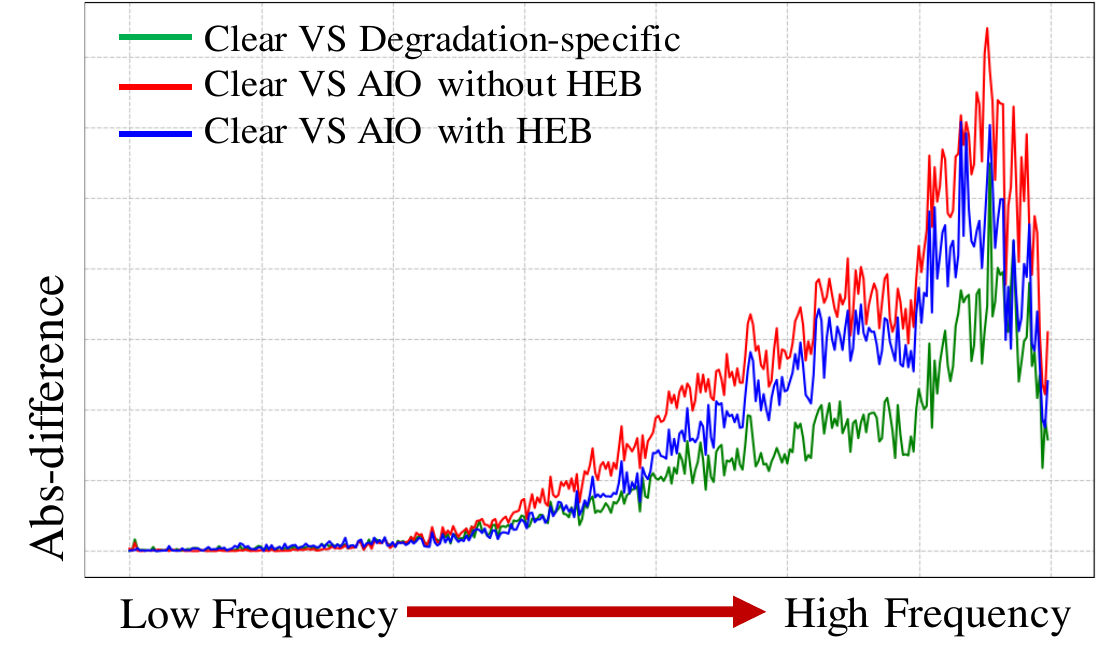}
    \caption{
    We compare frequency-domain differences between clear and restored images under three configurations. The All-in-One setting shows a notable high-frequency gap compared to the Degradation-specific setting, which is significantly reduced by adding the High-frequency Enhancement Block (HEB), as shown by the blue line.
    }
    \label{fig:fig5}
\end{figure}

\noindent\textbf{High-frequency Enhancement Block.}
\re{All-in-One image restoration necessitates the recovery of image components distributed across diverse frequency bands. While our proposed Degradation-aware Prompt State Space Model enables dynamic adaptation to degradation characteristics and partially mitigates inter-task conflicts, the joint optimization of heterogeneous degradations inevitably induces an implicit bias toward low-frequency structures. As illustrated in Fig.~\ref{fig:fig5}, this spectral bias results in suboptimal performance regarding high-frequency details, such as edges and textures, compared to degradation-specific models. To address this limitation, we introduce a lightweight High-Frequency Enhancement Block (HEB) as a complementary module to explicitly facilitate the restoration of local textures.}

For input features \(F\), we first extract channel-wise low-frequency features using global average pooling, then subtract them from \(F\) to obtain high-frequency components, as illustrated in Fig.~\ref{fig:fig4}(b). Finally, the high-frequency features are added back to enhance \(F\), formulated as:
\begin{equation}
    {F}_{enhanced} = {F} + \alpha \cdot \big({F} - \mathcal{G}({F})\big).
    \label{eq:11}
\end{equation}
Here, $\alpha$ represents a learnable scalar that controls the contribution of high-frequency enhancement, and $\mathcal{G}$ denotes the global average pooling operation. ${F}_{enhanced}$ is the enhanced feature map.

\subsection{Loss Function}
We employ the commonly used L1 loss $\mathcal{L}_1$ and L2 loss $\mathcal{L}_2$ in low-level vision tasks. Additionally, following previous works~\cite{zamir2022restormer, cui2023focal}, we incorporate a frequency-domain loss $\mathcal{L}_{fft}$ for training. The total loss is defined as:
\begin{equation}
    \mathcal{L}_{total} = \lambda_1 \cdot \mathcal{L}_1(O, \hat{O}) + \lambda_2 \cdot \mathcal{L}_2(O, \hat{O}) + \lambda_3 \cdot \mathcal{L}_{fft}(O, \hat{O}).
\end{equation}
where $\hat{O}$ and $O$ denote the model output and the ground truth, respectively. The parameters $\lambda_1$, $\lambda_2$, and $\lambda_3$ are balancing factors, which we set to 1, 0.5 and 0.001, respectively.

\begin{figure}
    \centering
    \includegraphics[width=0.9\linewidth]{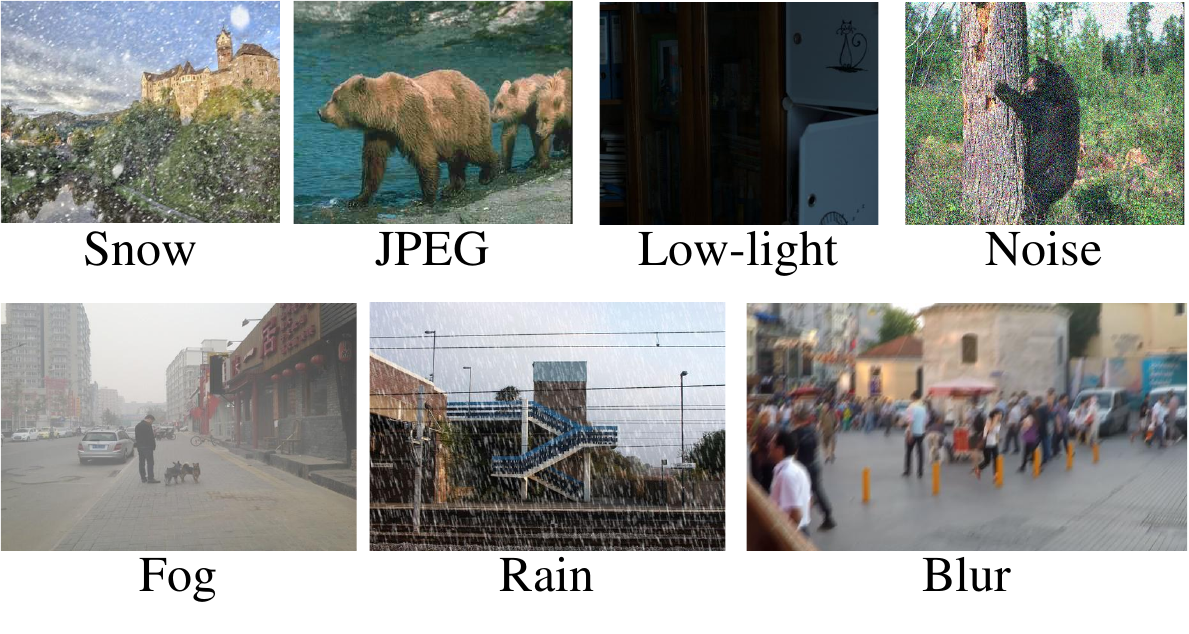}
    \caption{Examples of Seven Types of Image Degradations in the Dataset.}
    \label{fig:fig7}
\end{figure}

\begin{table*}
    \centering
    \renewcommand{\arraystretch}{1.3} 
    \setlength{\tabcolsep}{3pt} 
    
    \newcommand{\sep}{\,/\,}
    
    \caption{Quantitative Comparison on Seven Datasets for All-in-One Image Restoration. 
    The best and second-best results are shown in \textbf{bold} and \underline{underline}, respectively. 
    (Format: \textbf{PSNR} $\uparrow$ \sep \textbf{SSIM} $\uparrow$ \sep \textbf{LPIPS} $\downarrow$).}
    
    \adjustbox{max width=\textwidth}{
        \begin{tabular}{lcccccccc} 
            \toprule[1.5pt] 
            \textbf{Method} & \textbf{Lowlight} & \textbf{Snow} & \textbf{Rain} & \textbf{Haze} & \textbf{Jpeg} & \textbf{Noise} & \textbf{Blur} & \textbf{Average} \\ 
            \midrule[1pt] 
            
            MPRNet & 19.71\sep.659\sep.122 & 30.94\sep.938\sep.029 & 26.27\sep.912\sep.068 & 25.63\sep.947\sep.023 & 30.51\sep.938\sep.024 & 28.03\sep.897\sep.057 & 26.07\sep.869\sep.077 & 26.74\sep.880\sep.057 \\
            MIRNet & 19.81\sep.668\sep.123 & 31.04\sep.941\sep.028 & 25.77\sep.907\sep.073 & 22.26\sep.882\sep.058 & 30.54\sep.939\sep.024 & 28.14\sep.900\sep.057 & 26.32\sep.873\sep.070 & 26.27\sep.873\sep.062 \\
            NAFNet & 19.49\sep.650\sep.117 & 30.33\sep.935\sep.030 & 26.04\sep.906\sep.068 & 25.63\sep.946\sep.023 & 30.20\sep.936\sep.025 & 27.80\sep.895\sep{.051} & 26.13\sep.870\sep.075 & 26.52\sep.877\sep.056 \\
            Restormer & 19.99\sep.670\sep.113 & 31.67\sep.944\sep.025 & 26.91\sep.919\sep.061 & 26.18\sep.953\sep.020 & 30.56\sep.939\sep.024 & 28.17\sep.890\sep.056 & 26.62\sep.880\sep.063 & 27.16\sep.886\sep.052 \\
            MambaIR & 19.81\sep.664\sep.115 & 30.43\sep.935\sep.032 & 25.52\sep.901\sep.078 & 25.90\sep.944\sep.024 & 30.38\sep.937\sep.025 & 27.91\sep.895\sep.058 & 26.07\sep.869\sep.077 & 26.57\sep.878\sep.058 \\
            PromptIR & 19.90\sep.669\sep.116 & 31.52\sep.943\sep.024 & 26.80\sep.918\sep.061 & 26.13\sep.945\sep.022 & \underline{30.57}\sep\underline{.939}\sep.024 & 28.18\sep.901\sep.054 & 26.36\sep.874\sep.072 & 27.07\sep.884\sep.053 \\
            IDR & 19.68\sep.663\sep.115 & 31.16\sep.941\sep.027 & 26.39\sep.914\sep.065 & 26.11\sep.948\sep.023 & 30.51\sep.939\sep.024 & 28.10\sep.899\sep.054 & 26.29\sep.873\sep.073 & 26.89\sep.883\sep.054 \\
            NDR-Restore & 19.79\sep.635\sep.206 & 30.00\sep.902\sep.052 & 26.00\sep.813\sep.158 & 27.00\sep.943\sep.025 & 30.19\sep.872\sep.056 & 27.78\sep.786\sep.133 & 25.77\sep.791\sep.193 & 26.64\sep.820\sep.117 \\
            OneRestore & 19.77\sep.657\sep.127 & 29.95\sep.931\sep.034 & 25.37\sep.900\sep.080 & 27.74\sep.962\sep.016 & 30.10\sep.934\sep.026 & 27.65\sep.890\sep.060 & 25.90\sep.866\sep.082 & 26.64\sep.877\sep.061 \\
            MoCEIR & 19.92\sep.674\sep.109 & 31.22\sep.943\sep.025 & 26.56\sep.920\sep\textbf{.051} & 28.06\sep.963\sep\textbf{.014} & 30.51\sep.939\sep\textbf{.023} & 28.13\sep.900\sep{.053} & 26.34\sep.874\sep.069 & 27.25\sep.888\sep.049 \\
            AdaIR & 19.99\sep.672\sep\textbf{.102} & 31.94\sep.923\sep{.023} & 27.01\sep.922\sep.056 & 26.61\sep.952\sep.019 & 30.53\sep.939\sep.024 & 28.19\sep.900\sep.055 & 26.61\sep.879\sep.065 & 27.26\sep.887\sep.049 \\
            
            \midrule
            
            \rowcolor{gray!10}
            DPMambaIR & \underline{20.04}\sep\textbf{.680}\sep.109 & \textbf{32.36}\sep\underline{.949}\sep\underline{.022} & \underline{27.07}\sep\textbf{.926}\sep\underline{.055} & \underline{28.13}\sep\underline{.963}\sep.016 & \textbf{30.57}\sep\textbf{.939}\sep\underline{.024} & \textbf{28.25}\sep\textbf{.902}\sep\textbf{.051} & \underline{27.43}\sep\underline{.894}\sep\underline{.048} & \underline{27.69}\sep\underline{.893}\sep\underline{.047} \\ 
            \rowcolor{gray!10}
            DPMambaIR-L & \textbf{20.06}\sep\underline{.677}\sep\underline{.108} & \underline{32.31}\sep\textbf{.950}\sep\textbf{.022} & \textbf{27.11}\sep\underline{.925}\sep.056 & \textbf{28.39}\sep\textbf{.963}\sep\underline{.016} & 30.55\sep.939\sep{.024} & \underline{28.21}\sep\underline{.902}\sep\underline{.051} & \textbf{27.45}\sep\textbf{.895}\sep\textbf{.045} & \textbf{27.73}\sep\textbf{.893}\sep\textbf{.046} \\ 
            \bottomrule[1.5pt]
        \end{tabular}
    }
    \label{tab:tab2}
\end{table*}

\begin{table*}[t]
\centering
\caption{Quantitative Comparison on Deraining, Low-light Image Enhancement(LLIE), Deblurring and Dehazing for Degradation-Specific Image Restoration Task with State-of-the-Art Methods. The Best and Second-Best Results Are in \textbf{Bold} and \underline{Underline}.}
\label{tab:tab3}
\renewcommand{\arraystretch}{1.1} 
\begin{minipage}{0.24\textwidth}
\centering
\adjustbox{max width=\linewidth}{
\begin{tabular}{lcc}
\toprule
\textbf{Deraining} & PSNR & SSIM \\
\midrule
JORDER      & 26.25 & 0.835 \\
PReNet      & 29.46 & 0.899 \\
MPRNet      & 30.41 & 0.891 \\
MAXIM       & 30.81 & 0.903 \\
Restormer   & 31.46 & 0.904 \\
OneRestore  & 29.36 & 0.944 \\
MoCEIR      & 31.23 & 0.960 \\
AdaIR       & \underline{31.46} & \underline{0.961} \\
\rowcolor{gray!15}
DPMambaIR   & \textbf{32.30} & \textbf{0.967} \\
\bottomrule
\end{tabular}
}
\end{minipage}
\hfill
\begin{minipage}{0.24\textwidth}
\centering
\adjustbox{max width=\linewidth}{
\begin{tabular}{lcc}
\toprule
\textbf{LLIE} & PSNR & SSIM \\
\midrule
EnlightenGAN & 17.61 & 0.653 \\
MIRNet       & \underline{24.14} & 0.830 \\
URetinex-Net & 19.84 & 0.824 \\
MAXIM        & 23.43 & \textbf{0.863} \\
IAT          & 23.38 & 0.809 \\
OneRestore   & 22.97 & 0.835 \\
MoCEIR       & 23.64 & 0.818 \\
AdaIR        & 23.47 & 0.831 \\
\rowcolor{gray!15}
DPMambaIR    & \textbf{24.20} & \underline{0.852} \\
\bottomrule
\end{tabular}
}
\end{minipage}
\hfill
\begin{minipage}{0.24\textwidth}
\centering
\adjustbox{max width=\linewidth}{
\begin{tabular}{lcc}
\toprule
\textbf{Deblurring} & PSNR & SSIM \\
\midrule
DeepDeBlur  & 29.08 & 0.913 \\
DeBlurGAN   & 28.70 & 0.858 \\
DeBlurGANV2 & 29.55 & 0.934 \\
MT-RNN      & \underline{31.15} & \underline{0.945} \\
IR-SDE      & 30.70 & 0.901 \\
OneRestore  & 28.76 & 0.915 \\
MoCEIR      & 30.05 & 0.933 \\
AdaIR       & 30.57 & 0.939 \\
\rowcolor{gray!15}
DPMambaIR   & \textbf{31.42} & \textbf{0.948} \\
\bottomrule
\end{tabular}
}
\end{minipage}
\hfill
\begin{minipage}{0.24\textwidth}
\centering
\adjustbox{max width=\linewidth}{
\begin{tabular}{lcc}
\toprule
\textbf{Dehazing} & PSNR & SSIM \\
\midrule
GCANet        & 26.59 & 0.935 \\
GridDehazeNet & 25.86 & 0.944 \\
DeHazeFormer  & 30.29 & 0.964 \\
MAXIM         & 29.12 & 0.932 \\
DA-CLIP       & 30.16 & 0.936 \\
OneRestore    & 30.16 & 0.974 \\
MoCEIR        & 30.62 & 0.975 \\
AdaIR         & \underline{30.87} & \underline{0.975} \\
\rowcolor{gray!15}
DPMambaIR     & \textbf{30.89} & \textbf{0.977} \\
\bottomrule
\end{tabular}
}
\end{minipage}

\end{table*}

\section{Experiment and Analysis}
\noindent We evaluate our method on both All-in-one image restoration tasks and Degradation-specific image restoration tasks. In the All-in-One Image Restoration setting, we train a single model on a dataset containing seven types of degradation. For Degradation-specific image restoration, we train separately on datasets corresponding to each type of degradation.
\subsection{Experimental Settings}

\noindent\textbf{Training Details.} To ensure effective degradation-aware modeling, we adopt a \textbf{two-stage training strategy}. 
\re{In the first stage, we pre-train the Degradation Extractor using the same mixed-degradation dataset employed for the All-in-One task. The training pipeline consists of the Degradation Extractor, a Content Extractor, and a Degradation Reconstructor. Specifically, we utilize the pre-trained Restormer from DegAE~\cite{liu2023degae} as the Content Extractor with its parameters \textbf{frozen}, while optimizing only the Degradation Extractor and Reconstructor. This pre-training phase is conducted for 100,000 iterations with a batch size of 1, using input patches cropped to $256 \times 256$. We employ the AdamW~\cite{kingma2014adam} optimizer ($\beta_1 = 0.9$, $\beta_2 = 0.999$, weight decay $1e^{-4}$) with an initial learning rate of $1e^{-4}$, which is gradually reduced to $1e^{-6}$ through cosine annealing.}

\re{In the second stage, the pre-trained Degradation Extractor is integrated into the DPMambaIR framework with its parameters frozen to provide stable guidance.} 
Building on previous works~\cite{zamir2022restormer, promptir, guo2024mambair}, we train the main restoration network using the same AdamW configuration for 300,000 iterations. The initial learning rate is set to $3e^{-4}$ and decayed to $1e^{-6}$ via cosine annealing. Following~\cite{zamir2022restormer}, we employ a progressive training strategy, cropping image patches of size $192 \times 192$. Horizontal and vertical flips are applied for data augmentation. Our experiments are conducted on a single NVIDIA RTX A100 GPU and implemented using the PyTorch platform.

\noindent\textbf{Datasets.} For All-in-One image restoration, we collect a large-scale mixed dataset covering seven common degradation types: haze, rain, snow, low-light, blur, noise, and JPEG compression, as shown in Fig.~\ref{fig:fig7}. \zs{Detailed information regarding these datasets is summarized in Table~\ref{tab:datasets}.} Additionally, degradation-specific evaluations are conducted using LOL~\cite{lol} for low-light enhancement, GoPro~\cite{gopro} for deblurring, Rain100H~\cite{rain100H} for deraining, and RESIDE6K~\cite{reside6k} for dehazing.

\begin{table}[h!]
\centering
\caption{Datasets for Image Restoration Tasks}
\label{tab:datasets}
\begin{tabular}{l|lcc}
\hline
\textbf{Degradation} & \textbf{Dataset}       & \textbf{Train Pairs} & \textbf{Test Pairs} \\ \hline
Haze  & RESIDE6K~\cite{reside6k} & 6000                   & 1000                \\ 
Rain  & Rain13K~\cite{rain13k}   & 13711                 & -                   \\ 
      & Rain100H~\cite{rain100H} & -                      & 100                 \\
Snow  & Snow100K~\cite{snow100k} & 5000                   & 1000                \\ 
Low Light & LOL~\cite{lol}, LSRW~\cite{lsrw} & 6085          & 65               \\ 
Blur  & GoPro~\cite{gopro}       & 2103                   & 1111                \\ 
Noise & WaterlooED~\cite{waterlooed} & 4744 & - \\ 
      & BSD400~\cite{bsd400} & - & 400 \\ 
JPEG & WaterlooED~\cite{waterlooed}  & 4744 & - \\
      & BSD400~\cite{bsd400} & - & 400 \\   \hline
\end{tabular}
\end{table}

\begin{figure*}
    \centering
    \includegraphics[width=0.95\linewidth]{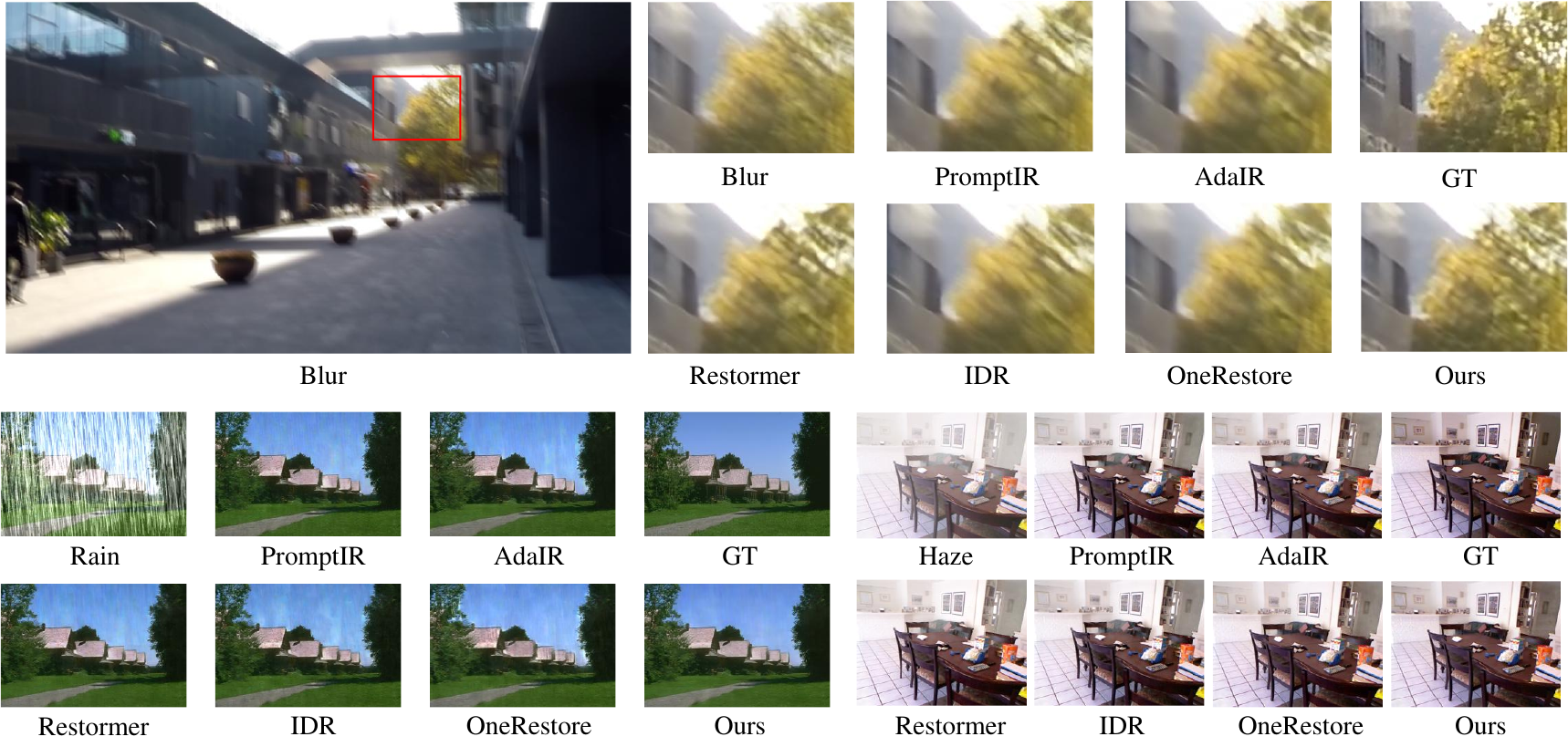}
    \caption{Qualitative comparison on Deblurring, Deraining and Dehazing for All-in-One Image Restoration Task.}
    \label{fig:fig6}
\end{figure*}

\noindent{\textbf{Evaluation Metrics.}}
\re{Following previous works~\cite{zamir2022restormer, guo2024mambair}, we employ Peak Signal-to-Noise Ratio (PSNR) and Structural Similarity Index Measure (SSIM) to quantitatively evaluate restoration quality on RGB channels. Additionally, we utilize the Learned Perceptual Image Patch Similarity (LPIPS) to assess the perceptual consistency of the restored images.}

\noindent\textbf{Comparisons with State-of-the-art Methods.}
For All-in-One image restoration, we compare against eleven methods: MIRNet~\cite{mirnet}, NAFNet~\cite{nafnet}, MPRNet~\cite{mprnet}, Restormer~\cite{zamir2022restormer}, MambaIR~\cite{guo2024mambair}, PromptIR~\cite{promptir}, IDR~\cite{idr}, NDR-Restore~\cite{ndrrestore}, OneRestore~\cite{guo2024onerestore}, MoCEIR~\cite{moceir}, and AdaIR~\cite{cui2025adair}. The first four are general-purpose restoration models, while the latter four are designed specifically for All-in-One restoration: PromptIR uses visual prompts, OneRestore employs a pre-trained degradation encoder with cross-attention, and AdaIR leverages frequency-domain information for adaptive restoration.
\zs{For degradation-specific restoration, we also include representative single-task methods such as JORDER~\cite{jorder} (deraining), MAXIM~\cite{maxim} (deblurring), EnlightenGAN~\cite{enlightengan} (low-light enhancement), DeepDeblur~\cite{deepdeblur}, AdaIR~\cite{cui2025adair}, DA-CLIP~\cite{daclip}, among others.}

\subsection{Quantitative Results}
\noindent
\textbf{All-in-one Image Restoration. }
We evaluated two versions of our method for comparison: the normal version (DPMambaIR) and a large version (DPMambaIR-L). \zs{The normal version adopts the degradation extractor illustrated in Fig.~\ref{fig:fig9}(c), while the large version uses the architecture shown in Fig.~\ref{fig:fig9}(b).} For subsequent evaluations, we primarily adopt the normal version unless otherwise specified. 
We compare our method against eleven state-of-the-art image restoration approaches. As shown in Table~\ref{tab:tab2}, for the All-in-One image restoration task, our method achieves the best PSNR and SSIM on the aforementioned mixed datasets, outperforming existing methods.
Specifically, our method achieves a PSNR of 27.69 dB and an SSIM of 0.893, surpassing AdaIR by 0.43 dB and 0.011, respectively. \zs{Additionally, DPMambaIR-L achieves a PSNR of 27.73 dB and an SSIM of 0.893, slightly outperforming the normal version.}

We achieve the best performance across all seven sub-tasks. Experimental results indicate that multi-task learning often encounters challenges in balancing performance across different tasks. For instance, AdaIR demonstrates strong performance in deraining but performs poorly in dehazing. On the other hand, OneRestore excels in dehazing but delivers suboptimal results in tasks such as deraining and deblurring. In contrast, our method effectively tackles the balance issue in multi-task learning, achieving consistently optimal performance across all sub-tasks. These findings highlight the effectiveness and superiority of our proposed method.

\noindent
\textbf{Degradation-Specific Image Restoration.}  
We evaluate our method on four Degradation-specific tasks, consistently achieving superior performance across multiple datasets compared to state-of-the-art CNN-based, transformer-based, and diffusion-based approaches, as shown in Table~\ref{tab:tab3}. On Rain100H \cite{rain100H}, our method achieves a PSNR of 32.30 dB and SSIM of 0.967, surpassing AdaIR by 0.84 dB and 0.006, respectively. On the LOL dataset \cite{lol}, it achieves 24.20 dB PSNR and 0.852 SSIM, outperforming MIRNet by 0.06 dB and 0.022. For deblurring on GoPro \cite{gopro}, it attains 31.42 dB PSNR and 0.948 SSIM, exceeding MT-RNN by 0.27 dB and 0.003. On RESIDE6K \cite{reside6k}, it achieves 30.89 dB PSNR and 0.973 SSIM, outperforming AdaIR by 0.02 dB and 0.002. These results highlight the versatility and effectiveness of our method in addressing diverse degradation restoration tasks.

\subsection{Qualitative Results}
Fig.~\ref{fig:fig6} presents the visual results of deblurring, deraining, and dehazing under the All-in-One Image Restoration task setting. Our method demonstrates superior visual performance compared to other approaches. Specifically, in deblurring, our method restores sharper edge details; in deraining, it more effectively removes rain streaks while recovering background information; and in dehazing, it focuses on regions often overlooked by other methods, resulting in more comprehensive dehazing performance. Additional visual comparisons under both the All-in-One and Degradation-specific settings are provided in the Appendix.

\subsection{\re{Model Complexity and Efficiency Analysis}}

\re{To accurately evaluate the practical applicability of DPMambaIR, we conducted a comprehensive efficiency comparison against state-of-the-art All-in-One methods. We measured the number of parameters, peak memory consumption, GFLOPs, and average inference time. All evaluations were performed on a single NVIDIA RTX A100 GPU with an input tensor size of $1 \times 3 \times 256 \times 256$. The statistics for DPMambaIR include the overhead of the Degradation Extractor.}

\re{As summarized in Table \ref{tab:efficiency}, DPMambaIR achieves the best performance while maintaining comparable computational overhead. Compared to the second-best method, AdaIR, our model delivers a significant performance gain of 0.43 dB with similar parameters and GFLOPs. }

\begin{table}[t]
    \caption{Comparison of complexity and efficiency against SOTA methods. Metrics: Parameters (M), Peak Memory (MB), GFLOPs, and Inference Time (ms) on $256 \times 256$ inputs.}
    \label{tab:efficiency}
    \centering
    \adjustbox{max width=\linewidth}{
        \begin{tabular}{l|ccccc}
        \toprule
        Method & \makecell[c]{Params\\(M)} & \makecell[c]{Mem.\\(MB)} & \makecell[c]{FLOPs\\(G)} & \makecell[c]{Inf Time\\(ms)} & \makecell[c]{PSNR\\(dB)} \\
        \midrule
        Restormer & 26.13 & 676.24 & 154.88 & 84.283 & 27.16 \\
        IDR      & 36.28 & 1433.24 & 270.76 & 81.977 & 26.89 \\
        PromptIR & 35.59 & 720.10 & 172.71 & 90.964 & 27.07 \\
        MoCEIR   & 25.35 & 404.48 & 97.74 & 93.232 & 27.25 \\
        AdaIR    & 28.78 & 686.40 & 161.76 & 102.903 & 27.26 \\
        \rowcolor{gray!20}
        \textbf{DPMambaIR} & 29.35 & 718.76 & 153.33 & 90.869 & \textbf{27.69} \\
        \bottomrule
        \end{tabular}
    }
\end{table}

\subsection{Ablation Study}
We conduct ablation studies to evaluate our key designs: the degradation-aware prompts and High-frequency Enhancement Block (HEB) (Table~\ref{tab:ablation_core}), degradation extraction and utilization methods (Table~\ref{tab:tab5}), degradation embedding dimension (Table~\ref{tab:ablation_dim}) and the global information supplementation strategy (Table~\ref{tab:tab6}).

Table~\ref{tab:ablation_core} demonstrates the effectiveness of our proposed core modules.
First, the High-frequency Enhancement Block (HEB) alone improves the baseline by 0.29 dB (Model (a)), validating its efficacy in enhancing local details.
Second, to verify the importance of the degradation-aware mechanism, we conduct a detailed breakdown of the prompt components.
Notably, employing $P_\Delta$ alone (Model (c)) achieves 27.58 dB, which slightly outperforms the combination of $P_B$ and $P_C$ (Model (b), 27.56 dB).
This observation underscores that dynamically modulating the discretization step $\Delta$, which fundamentally alters the state transition dynamics, is the most critical factor for adapting to diverse degradation types.
Furthermore, combining $P_\Delta$ with $P_B$ or $P_C$ yields consistent improvements, and integrating all prompt modules (Model (f)) pushes the PSNR to 27.67 dB.
Finally, the full model (Model (g)) demonstrates the synergy between prompt-based modulation and frequency enhancement, achieving the best performance of 27.69 dB.

\begin{figure}[t]
    \centering
    \includegraphics[width=0.8\linewidth]{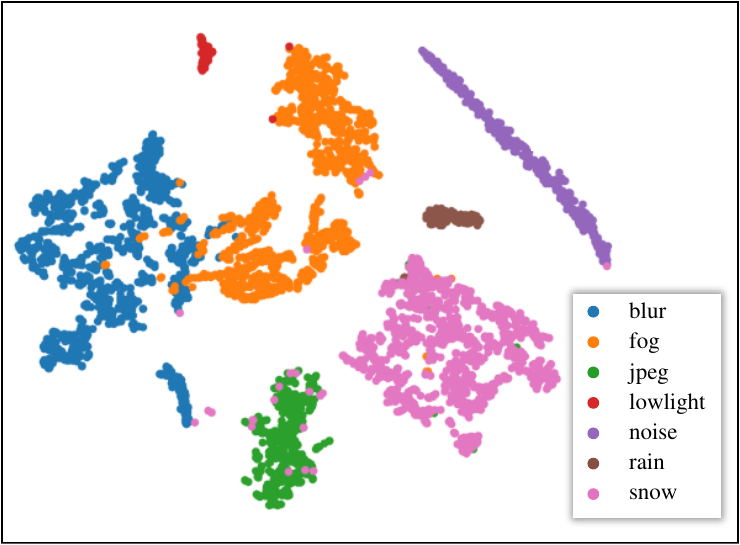}
    \caption{t-SNE visualization of degradation embeddings.}
    \label{fig:fig8}
\end{figure}

Table~\ref{tab:tab5} presents the results for degradation extraction and utilization methods. For degradation extraction, we compare three approaches: (1) “Classes”, which provides manually specified degradation types and can be regarded as a 100\% accurate non-blind setting; (2) a pre-trained degradation classifier from OneRestore~\cite{guo2024onerestore}; and (3) our fine-grained degradation representation via image reconstruction. 
The OneRestore extractor, optimized for classification, is insensitive to degradation strength or location, leading to poorer results.
Our method surpasses the non-blind "Classes" baseline by 0.11 dB PSNR.
Fig.~\ref{fig:fig8} visualizes the t-SNE embeddings of the extracted degradation features.
Significant gaps exist between different degradation types, indicating clear separability. Notably, although JPEG and noise share the same content during synthesis, the t-SNE visualization reveals that their separability is primarily driven by degradation information rather than image content.

For degradation utilization, we compare concatenation and attention-based fusion with our prompt modulation design. 
Our prompt-based modulation outperforms simple concatenation or attention-based fusion, showing better adaptability to local degradations.

\re{For the training strategy, compared with jointly fine-tuning the degradation extractor together with the backbone network, freezing the parameters of the degradation extractor provides a more stable supply of degradation information. This design choice leads to more consistent guidance during restoration and results in a clear performance gain of 0.26 dB in PSNR, demonstrating the advantage of decoupling degradation representation learning from the restoration optimization process.}

\begin{table}[t]
    \centering
    \caption{\re{Extended ablation studies on the proposed core modules. We investigate the individual and combined contributions of the prompt components ($P_\Delta, P_B, P_C$) and the High-frequency Enhancement Block (HEB).}}
    \adjustbox{max width=\linewidth}{
        \begin{tabular}{lcccc|cc}
        \toprule
        \textbf{Method} & \(P_\Delta\) & \(P_B\) & \(P_C\) & HEB & PSNR & SSIM \\ 
        \midrule
        Baseline & \usym{2717} & \usym{2717} & \usym{2717} & \usym{2717} & 26.96 & 0.884 \\
        (a) & \usym{2717} & \usym{2717} & \usym{2717} & \usym{2713} & 27.25 & 0.887 \\
        \midrule
        (b) & \usym{2717} & \usym{2713} & \usym{2713} & \usym{2717} & 27.56 & 0.891 \\
        (c) & \usym{2713} & \usym{2717} & \usym{2717} & \usym{2717} & 27.58 & 0.892 \\
        (d) & \usym{2713} & \usym{2713} & \usym{2717} & \usym{2717} & 27.60 & 0.892 \\
        (e) & \usym{2713} & \usym{2717} & \usym{2713} & \usym{2717} & 27.61 & 0.892 \\
        (f) & \usym{2713} & \usym{2713} & \usym{2713} & \usym{2717} & 27.67 & 0.893 \\
        \midrule
        \rowcolor{gray!15}
        \textbf{(g)} & \usym{2713} & \usym{2713} & \usym{2713} & \usym{2713} & \textbf{27.69} & \textbf{0.893} \\
        \bottomrule
        \end{tabular}
    }
    \label{tab:ablation_core}
\end{table}

\begin{table}[t]
    \centering
    \caption{Ablation Study on Different Degradation Extraction and Utilization Methods.}
    \label{tab:tab5}
    
    \adjustbox{max width=\linewidth}{
        \begin{tabular}{llcc}
            \toprule
            \textbf{Module} & \textbf{Method} & \textbf{PSNR} & \textbf{SSIM} \\ 
            \midrule
            \multirow{3}{*}{Degradation Extraction} 
                & Classes        & 27.58 & 0.892 \\
                & OneRestore*    & 27.56 & 0.891 \\
                & \textbf{Ours}  & \textbf{27.69} & \textbf{0.893} \\ 
            \midrule
            \multirow{3}{*}{Degradation Insertion} 
                & Concat         & 26.96 & 0.882 \\
                & Attention      & 27.39 & 0.882 \\
                & \textbf{Modulation (Ours)}  & \textbf{27.69} & \textbf{0.893} \\ 
            \midrule
            \multirow{2}{*}{Training Strategy} 
                & Fine-tuning    & 27.43 & 0.888 \\
                & \textbf{Frozen (Ours)} & \textbf{27.69} & \textbf{0.893} \\ 
            \bottomrule
        \end{tabular}
    }
\end{table}

\begin{table}[t]
    \centering
    \caption{\re{Ablation study on the dimension of degradation embedding ($D_{emb}$). We evaluate model complexity (Params, GFLOPs), auxiliary task performance (Rec. PSNR, Cls. Acc.), and final All-in-One restoration quality (AIO PSNR). The default configuration is gray.}}
    \label{tab:ablation_dim}
    \adjustbox{max width=\linewidth}{
        \begin{tabular}{c|cc|cc|c}
            \toprule
            \multirow{2}{*}{$D_{emb}$} & Params & FLOPs & Rec. PSNR & Cls. Acc. & AIO. PSNR \\
             & (M) & (G) & (dB) & (\%) & (dB) \\
            \midrule
            256  & 28.71 & 153.33 & 21.51 & 72.0 & 27.55 \\
            384  & 29.03 & 153.33 & 24.53 & 84.3 & 27.63 \\
            \rowcolor{gray!20} 
            512  & 29.35 & 153.33 & 25.67 & 92.0 & 27.69 \\
            1024 & 30.62 & 153.33 & 25.78 & 92.6 & 27.69 \\
            \bottomrule
        \end{tabular}
    }
\end{table}

\begin{table}[t]
    \centering
    \caption{Ablation Study on Different Methods for Global Information Supplementation.}
    \label{tab:tab6}
    \renewcommand{\arraystretch}{1.2} 

    \adjustbox{max width=\linewidth}{
        \begin{tabular}{lcccc}
        \toprule
        \textbf{Method} & \textbf{PSNR} & \textbf{SSIM} & \textbf{Params (M)} & \textbf{GFLOPs} \\ 
        \midrule
        Bi-direction Scan       & 27.66 & 0.893 & 31.43 & 244.3 \\ 
        \rowcolor{gray!15}
        \textbf{Prompt-C (Ours)}  & \textbf{27.69} & \textbf{0.893} & \textbf{29.35} & \textbf{153.3} \\ 
        \bottomrule
        \end{tabular}
    }
\end{table}


\re{Table~\ref{tab:ablation_dim} presents the ablation study on the Degradation Embedding dimension ($D_{emb}$). Given that the Degradation Extractor serves as the sole source of degradation priors, the completeness and distinctiveness of the extracted information are paramount. To quantitatively evaluate the quality of the learned embeddings, we introduce two direct indicators: (1) \textbf{Degradation Reconstruction PSNR (Rec. PSNR)}, calculated between the reconstructed degraded image and the target degraded input, which measures information integrity; and (2) \textbf{Classification Accuracy (Cls. Acc.)}, which assesses the semantic separability of the degradation types.}
\re{For the classification metric, we appended a simple MLP head to the frozen degradation extractor and trained it for only 1,000 iterations to map the embeddings to the seven degradation classes. As observed in Table~\ref{tab:ablation_dim}, increasing $D_{emb}$ from 256 to 512 yields substantial gains: Rec. PSNR improves from 21.51 dB to 25.67 dB, and Cls. Acc. surges from 72.0\% to 92.0\%. This indicates that higher-dimensional embeddings effectively encode both the pixel-level details and the semantic categories of degradations. These improvements directly translate to the final restoration task, boosting AIO. PSNR to 27.69 dB. However, expanding to 1024 yields marginal returns despite higher costs. Thus, 512 is selected as the optimal balance.}

Table~\ref{tab:tab6} presents the ablation study on global information supplementation. We compare our Prompt-C with Mamba's bi-directional scanning scheme. Compared to Mamba's bi-directional scan, our Prompt-C not only achieves slightly better performance but is also more efficient, reducing GFLOPs by nearly 37\%. This highlights its effectiveness as a lightweight global information supplementation method.

\subsection{Generalization Capability and Failure Boundary Analysis}
\re{To investigate the robustness of DPMambaIR beyond the training distribution, we conducted Out-of-Distribution (OOD) evaluations on two unseen degradation types: Raindrop and Pixelation. Importantly, these evaluations were performed by directly applying the pre-trained model to the test images without any additional training or fine-tuning. This rigorous setting serves to delineate the generalization boundaries of our Degradation-Aware Prompt mechanism.}

\re{\textbf{Robustness on Unseen Degradations.} 
We evaluated the model on the Raindrop dataset\cite{raindrop} (58 images) and a synthetically generated Pixelated BSD68 dataset\cite{bsd68} (block size 2). As summarized in Table \ref{tab:ood_results}, DPMambaIR exhibits superior generalization, securing top-tier performance across all metrics. Specifically, on the Pixelation task, it achieves a PSNR of 23.35 dB, surpassing the second-best AdaIR by a significant margin of 1.96 dB.
We attribute this robustness to our regression-based embedding strategy. Unlike static prompt approaches that risk model collapse on undefined inputs, DPMambaIR projects unseen degradations into a continuous latent space. This mechanism enables the model to approximate unknown corruption patterns within a learned manifold, preventing inference failure and yielding visually plausible results.}
\re{A notable phenomenon is further observed in a specific low-light sample from the BSD68 dataset. When processing the pixelated version of this image, DPMambaIR automatically performs brightness enhancement alongside depixelation, capturing and correcting the underlying low-light degradation. In contrast, competing methods fail to address this compound issue. This capability underscores the potential of DPMambaIR in handling mixed degradations, suggesting that future work can further explore real-world restoration via more complex, fine-grained coupled degradation modeling.}

\begin{figure}[t]
    \centering
    \includegraphics[width=\linewidth]{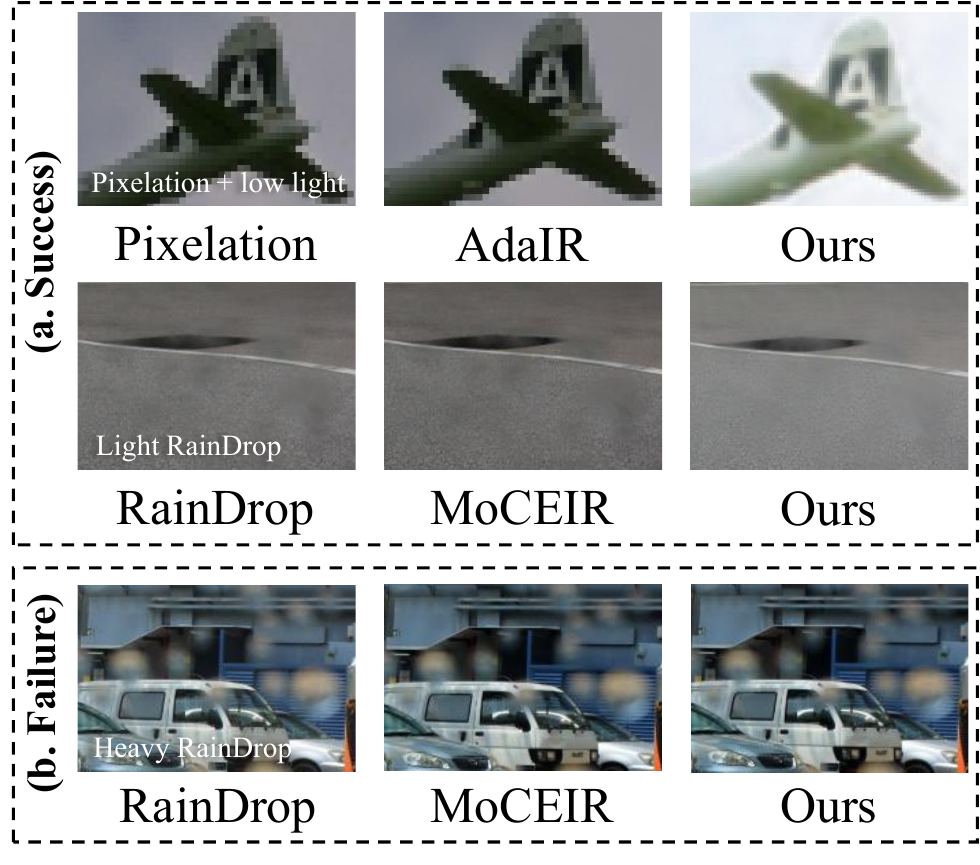} 
    \caption{\re{Visual analysis of generalization and failure boundaries. \textbf{(a) Success Cases:} The top row demonstrates restoration on unseen Pixelation, where DPMambaIR concurrently removes artifacts and enhances inherent low-light details. The middle row shows effective removal of Light Raindrop. \textbf{(b) Failure Case:} The bottom row illustrates the boundary on Heavy Raindrop, where the model fails to hallucinate the background behind opaque occlusions.}}
    \label{fig:ood_analysis}
\end{figure}

\re{\textbf{Failure Boundary Analysis.} 
Despite these promising results, DPMambaIR is not without limitations. We observed a distinct performance bottleneck when processing images with heavy raindrops. As visually analyzed in Fig. \ref{fig:ood_analysis}, while the model effectively removes light raindrops, it struggles with large, opaque water droplets that completely occlude the background. 
We attribute this to the physical nature of the degradation: removing heavy, opaque occlusions requires generative hallucination (inpainting) to synthesize missing content, whereas DPMambaIR operates on restoration cues. Future work could explore incorporating generative priors, such as Diffusion Models, to address such severe occlusions.}

\begin{table}[t]
    \caption{\re{Quantitative comparison on unseen degradation types (OOD Testing). The best and second-best results are highlighted in \textbf{bold} and \underline{underline}.}}
    \label{tab:ood_results}
    \centering
    \renewcommand{\arraystretch}{1.2}
    \setlength{\tabcolsep}{2mm}
    \footnotesize
    \begin{tabular}{lcc}
        \toprule
        \multirow{2}{*}{\textbf{Method}} & \textbf{Pixelation} & \textbf{Raindrop} \\
         & (PSNR / SSIM / LPIPS) & (PSNR / SSIM / LPIPS) \\
        \midrule
        MambaIR    & 20.80 / 0.7601 / 0.0568 & 19.29 / 0.7992 / 0.1051 \\
        PromptIR   & 19.69 / 0.7479 / 0.0582 & 19.26 / 0.7775 / 0.1146 \\
        IDR        & 20.64 / 0.7567 / \underline{0.0494} & 17.40 / 0.7609 / 0.1276 \\
        OneRestore & 19.75 / 0.7415 / 0.0628 & 19.75 / 0.7792 / 0.1100 \\
        AdaIR      & \underline{21.39} / \underline{0.7676} / 0.0512 & 19.07 / 0.7722 / 0.1225 \\
        MoCEIR     & 19.89 / 0.7355 / 0.0544 & \underline{21.46} / \underline{0.8254} / \underline{0.0919} \\
        \midrule
        \textbf{DPMambaIR} & \textbf{23.35} / \textbf{0.8003} / \textbf{0.0551} & \textbf{21.63} / \textbf{0.8337} / \textbf{0.0932} \\
        \bottomrule
    \end{tabular}
\end{table}


\section{Conclusion}
We propose DPMambaIR, a novel All-in-One framework for image restoration capable of handling diverse degradation types. The core of our framework is a Degradation-Aware Prompt State Space Model (DP-SSM) that leverages a fine-grained degradation extractor. This design enables the dynamic integration of degradation features into the state-space modeling process, allowing for adaptive handling of complex degradation scenarios while maintaining global degradation awareness. Furthermore, a lightweight High-frequency Enhancement Block (HEB) is introduced to complement the main framework by enhancing high-frequency detail restoration with negligible computational overhead. This study underscores the importance of fine-grained degradation modeling and dynamic feature modulation in advancing All-in-One image restoration frameworks. Extensive experiments on a mixed dataset containing seven degradation types show that DPMambaIR achieves the best performance, establishing it as a promising approach for unified image restoration.

\bibliographystyle{IEEEtran}
\bibliography{sample-base}

@article{zhang2021deep,
  title={Deep dense multi-scale network for snow removal using semantic and depth priors},
  author={Zhang, Kaihao and Li, Rongqing and Yu, Yanjiang and Luo, Wenhan and Li, Changsheng},
  journal={IEEE Transactions on Image Processing},
  volume={30},
  pages={7419--7431},
  year={2021},
  publisher={IEEE}
}

@inproceedings{raindrop,
  title={Attentive generative adversarial network for raindrop removal from a single image},
  author={Qian, Rui and Tan, Robby T and Yang, Wenhan and Su, Jiajun and Liu, Jiaying},
  booktitle={Proceedings of the IEEE conference on computer vision and pattern recognition},
  pages={2482--2491},
  year={2018}
}

@inproceedings{bsd68,
  title={A database of human segmented natural images and its application to evaluating segmentation algorithms and measuring ecological statistics},
  author={Martin, David and Fowlkes, Charless and Tal, Doron and Malik, Jitendra},
  booktitle={Proceedings eighth IEEE international conference on computer vision. ICCV 2001},
  volume={2},
  pages={416--423},
  year={2001},
  organization={IEEE}
}

@inproceedings{liu2023degae,
  title={Degae: A new pretraining paradigm for low-level vision},
  author={Liu, Yihao and He, Jingwen and Gu, Jinjin and Kong, Xiangtao and Qiao, Yu and Dong, Chao},
  booktitle={Proceedings of the IEEE/CVF Conference on Computer Vision and Pattern Recognition},
  pages={23292--23303},
  year={2023}
}

@article{zhang2018adversarial,
  title={Adversarial spatio-temporal learning for video deblurring},
  author={Zhang, Kaihao and Luo, Wenhan and Zhong, Yiran and Ma, Lin and Liu, Wei and Li, Hongdong},
  journal={IEEE Transactions on Image Processing},
  volume={28},
  number={1},
  pages={291--301},
  year={2018},
  publisher={IEEE}
}

@article{zhang2023mc,
  title={MC-Blur: A comprehensive benchmark for image deblurring},
  author={Zhang, Kaihao and Wang, Tao and Luo, Wenhan and Ren, Wenqi and Stenger, Bj{\"o}rn and Liu, Wei and Li, Hongdong and Yang, Ming-Hsuan},
  journal={IEEE Transactions on Circuits and Systems for Video Technology},
  volume={34},
  number={5},
  pages={3755--3767},
  year={2023},
  publisher={IEEE}
}

@article{zhang2022enhanced,
  title={Enhanced spatio-temporal interaction learning for video deraining: faster and better},
  author={Zhang, Kaihao and Li, Dongxu and Luo, Wenhan and Ren, Wenqi and Liu, Wei},
  journal={IEEE Transactions on Pattern Analysis and Machine Intelligence},
  volume={45},
  number={1},
  pages={1287--1293},
  year={2022},
  publisher={IEEE}
}

@article{jin2025mb,
  title={MB-TaylorFormer V2: improved multi-branch linear transformer expanded by Taylor formula for image restoration},
  author={Jin, Zhi and Qiu, Yuwei and Zhang, Kaihao and Li, Hongdong and Luo, Wenhan},
  journal={IEEE Transactions on Pattern Analysis and Machine Intelligence},
  year={2025},
  publisher={IEEE}
}

@ARTICLE{tmmsr1,
  author={Liu, Shuzhe and Suzhang, Delong and Yang, Meng and Zheng, Xinhu and Zhu, Ce},
  journal={IEEE Transactions on Multimedia}, 
  title={Depth Map Super-Resolution via Deep Cross-modality and Cross-scale Guidance}, 
  year={2025},
  volume={},
  number={},
  pages={1-14},
  doi={10.1109/TMM.2025.3607763}
}

@ARTICLE{tmmdenoising1,
  author={Zhang, Ting and Wang, Runjie and Niu, Yuzhen and Li, Zuoyong and Zhao, Tiesong},
  journal={IEEE Transactions on Multimedia}, 
  title={HUGS-Net: A Lightweight and Unified Network for Adverse Weather Image Denoising}, 
  year={2025},
  volume={},
  number={},
  pages={1-10},
  doi={10.1109/TMM.2025.3613104}}

@ARTICLE{tmmdeblur1,
  author={Zhuang, Kai and Li, Qiang and Yuan, Yuan and Wang, Qi},
  journal={IEEE Transactions on Multimedia}, 
  title={Multi-Domain Adaptation for Motion Deblurring}, 
  year={2024},
  volume={26},
  number={},
  pages={3676-3688},
  doi={10.1109/TMM.2023.3314154}}

@ARTICLE{tmmderain1,
  author={Peng, Yan-Tsung and Li, Wei-Hua and Chen, Zihao},
  journal={IEEE Transactions on Multimedia}, 
  title={Rain2Avoid: Learning Deraining by Self-Supervision}, 
  year={2025},
  volume={27},
  number={},
  pages={4765-4779},
  doi={10.1109/TMM.2025.3542981}}

@ARTICLE{tmmdehaze,
  author={Su, Yanzhao and Wang, Nian and Cui, Zhigao and Cai, Yanping and He, Chuan and Li, Aihua},
  journal={IEEE Transactions on Multimedia}, 
  title={Real Scene Single Image Dehazing Network With Multi-Prior Guidance and Domain Transfer}, 
  year={2025},
  volume={27},
  number={},
  pages={5492-5506},
  doi={10.1109/TMM.2025.3543063}}

@ARTICLE{tmmlol,
  author={Morawski, Igor and He, Kai and Dangi, Shusil and Hsu, Winston H.},
  journal={IEEE Transactions on Multimedia}, 
  title={Leveraging Content and Context Cues for Low-Light Image Enhancement}, 
  year={2025},
  volume={27},
  number={},
  pages={5337-5351},
  doi={10.1109/TMM.2025.3543047}}

@article{bsd400,
  title={Contour detection and hierarchical image segmentation},
  author={Arbelaez, Pablo and Maire, Michael and Fowlkes, Charless and Malik, Jitendra},
  journal={IEEE transactions on pattern analysis and machine intelligence},
  volume={33},
  number={5},
  pages={898--916},
  year={2010},
  publisher={IEEE}
}

@article{snow100k,
  title={Desnownet: Context-aware deep network for snow removal},
  author={Liu, Yun-Fu and Jaw, Da-Wei and Huang, Shih-Chia and Hwang, Jenq-Neng},
  journal={IEEE Transactions on Image Processing},
  volume={27},
  number={6},
  pages={3064--3073},
  year={2018},
  publisher={IEEE}
}

@article{waterlooed,
  title={Waterloo exploration database: New challenges for image quality assessment models},
  author={Ma, Kede and Duanmu, Zhengfang and Wu, Qingbo and Wang, Zhou and Yong, Hongwei and Li, Hongliang and Zhang, Lei},
  journal={IEEE Transactions on Image Processing},
  volume={26},
  number={2},
  pages={1004--1016},
  year={2016},
  publisher={IEEE}
}

@article{lsrw,
  title={R2rnet: Low-light image enhancement via real-low to real-normal network},
  author={Hai, Jiang and Xuan, Zhu and Yang, Ren and Hao, Yutong and Zou, Fengzhu and Lin, Fang and Han, Songchen},
  journal={Journal of Visual Communication and Image Representation},
  volume={90},
  pages={103712},
  year={2023}
}

@inproceedings{rain13k,
  title={Multi-scale progressive fusion network for single image deraining},
  author={Jiang, Kui and Wang, Zhongyuan and Yi, Peng and Chen, Chen and Huang, Baojin and Luo, Yimin and Ma, Jiayi and Jiang, Junjun},
  booktitle={Proceedings of the IEEE/CVF conference on computer vision and pattern recognition},
  pages={8346--8355},
  year={2020}
}

@article{ndrrestore,
  title={Neural degradation representation learning for all-in-one image restoration},
  author={Yao, Mingde and Xu, Ruikang and Guan, Yuanshen and Huang, Jie and Xiong, Zhiwei},
  journal={IEEE Transactions on Image Processing},
  year={2024},
  publisher={IEEE}
}

@misc{moceir,
      title={Complexity Experts are Task-Discriminative Learners for Any Image Restoration}, 
      author={Eduard Zamfir and Zongwei Wu and Nancy Mehta and Yuedong Tan and Danda Pani Paudel and Yulun Zhang and Radu Timofte},
      year={2024},
      eprint={2411.18466},
      archivePrefix={arXiv},
      primaryClass={cs.CV},
}

@inproceedings{guo2024mambair,
  title={Mambair: A simple baseline for image restoration with state-space model},
  author={Guo, Hang and Li, Jinmin and Dai, Tao and Ouyang, Zhihao and Ren, Xudong and Xia, Shu-Tao},
  booktitle={European conference on computer vision},
  pages={222--241},
  year={2024},
  organization={Springer}
}

@inproceedings{zhang2024MoFME,
  title={Efficient deweahter mixture-of-experts with uncertainty-aware feature-wise linear modulation},
  author={Zhang, Rongyu and Luo, Yulin and Liu, Jiaming and Yang, Huanrui and Dong, Zhen and Gudovskiy, Denis and Okuno, Tomoyuki and Nakata, Yohei and Keutzer, Kurt and Du, Yuan and others},
  booktitle={Proceedings of the AAAI Conference on Artificial Intelligence},
  volume={38},
  number={15},
  pages={16812--16820},
  year={2024}
}

@inproceedings{guo2024onerestore,
  title={Onerestore: A universal restoration framework for composite degradation},
  author={Guo, Yu and Gao, Yuan and Lu, Yuxu and Zhu, Huilin and Liu, Ryan Wen and He, Shengfeng},
  booktitle={European Conference on Computer Vision},
  pages={255--272},
  year={2024},
  organization={Springer}
}

@article{he2010single,
  title={Single image haze removal using dark channel prior},
  author={He, Kaiming and Sun, Jian and Tang, Xiaoou},
  journal={IEEE transactions on pattern analysis and machine intelligence},
  volume={33},
  number={12},
  pages={2341--2353},
  year={2010},
  publisher={IEEE}
}

@article{fattal2014dehazing,
  title={Dehazing using color-lines},
  author={Fattal, Raanan},
  journal={ACM transactions on graphics (TOG)},
  volume={34},
  number={1},
  pages={1--14},
  year={2014},
  publisher={ACM New York, NY, USA}
}

@inproceedings{kim2024lan,
  title={Lan: Learning to adapt noise for image denoising},
  author={Kim, Changjin and Kim, Tae Hyun and Baik, Sungyong},
  booktitle={Proceedings of the IEEE/CVF Conference on Computer Vision and Pattern Recognition},
  pages={25193--25202},
  year={2024}
}

@inproceedings{wang2024odcr,
  title={ODCR: Orthogonal Decoupling Contrastive regularization for unpaired image Dehazing},
  author={Wang, Zhongze and Zhao, Haitao and Peng, Jingchao and Yao, Lujian and Zhao, Kaijie},
  booktitle={Proceedings of the IEEE/CVF Conference on Computer Vision and Pattern Recognition},
  pages={25479--25489},
  year={2024}
}

@article{guo2025deep,
  title={Deep Unfolding Network for Image Desnowing with Snow Shape Prior},
  author={Guo, Xin and Wang, Xi and Fu, Xueyang and Zha, Zheng-Jun},
  journal={IEEE Transactions on Circuits and Systems for Video Technology},
  year={2025},
  publisher={IEEE}
}

@article{lecun2015deep,
  title={Deep learning},
  author={LeCun, Yann and Bengio, Yoshua and Hinton, Geoffrey},
  journal={nature},
  volume={521},
  number={7553},
  pages={436--444},
  year={2015},
  publisher={Nature Publishing Group UK London}
}

@article{lecun1989handwritten,
  title={Handwritten digit recognition with a back-propagation network},
  author={LeCun, Yann and Boser, Bernhard and Denker, John and Henderson, Donnie and Howard, Richard and Hubbard, Wayne and Jackel, Lawrence},
  journal={Advances in neural information processing systems},
  volume={2},
  year={1989}
}

@article{vaswani2017attention,
  title={Attention is all you need},
  author={Vaswani, Ashish and Shazeer, Noam and Parmar, Niki and Uszkoreit, Jakob and Jones, Llion and Gomez, Aidan N and Kaiser, {\L}ukasz and Polosukhin, Illia},
  journal={Advances in neural information processing systems},
  volume={30},
  year={2017}
}

@article{ho2020denoising,
  title={Denoising diffusion probabilistic models},
  author={Ho, Jonathan and Jain, Ajay and Abbeel, Pieter},
  journal={Advances in neural information processing systems},
  volume={33},
  pages={6840--6851},
  year={2020}
}

@inproceedings{airnet,
  title={All-in-one image restoration for unknown corruption},
  author={Li, Boyun and Liu, Xiao and Hu, Peng and Wu, Zhongqin and Lv, Jiancheng and Peng, Xi},
  booktitle={Proceedings of the IEEE/CVF conference on computer vision and pattern recognition},
  pages={17452--17462},
  year={2022}
}

@article{promptir,
  title={Promptir: Prompting for all-in-one image restoration},
  author={Potlapalli, Vaishnav and Zamir, Syed Waqas and Khan, Salman H and Shahbaz Khan, Fahad},
  journal={Advances in Neural Information Processing Systems},
  volume={36},
  pages={71275--71293},
  year={2023}
}

@inproceedings{idr,
  title={Ingredient-oriented multi-degradation learning for image restoration},
  author={Zhang, Jinghao and Huang, Jie and Yao, Mingde and Yang, Zizheng and Yu, Hu and Zhou, Man and Zhao, Feng},
  booktitle={Proceedings of the IEEE/CVF conference on computer vision and pattern recognition},
  pages={5825--5835},
  year={2023}
}

@inproceedings{cui2025adair,
title={Ada{IR}: Adaptive All-in-One Image Restoration via Frequency Mining and Modulation},
author={Yuning Cui and Syed Waqas Zamir and Salman Khan and Alois Knoll and Mubarak Shah and Fahad Shahbaz Khan},
booktitle={The Thirteenth International Conference on Learning Representations},
year={2025}
}

@article{daclip,
  title={Controlling Vision-Language Models for Universal Image Restoration},
  author={Luo, Ziwei and Gustafsson, Fredrik K and Zhao, Zheng and Sj{\"o}lund, Jens and Sch{\"o}n, Thomas B},
  journal={arXiv preprint arXiv:2310.01018},
  year={2023}
}

@article{vim,
  title={Vision Mamba: Efficient Visual Representation Learning with Bidirectional State Space Model},
  author={Lianghui Zhu and Bencheng Liao and Qian Zhang and Xinlong Wang and Wenyu Liu and Xinggang Wang},
  journal={arXiv preprint arXiv:2401.09417},
  year={2024}
}

@article{kalman1960new,
  title={A new approach to linear filtering and prediction problems},
  author={Kalman, Rudolph Emil},
  year={1960}
}

@article{vit,
  title={An image is worth 16x16 words: Transformers for image recognition at scale},
  author={Dosovitskiy, Alexey and Beyer, Lucas and Kolesnikov, Alexander and Weissenborn, Dirk and Zhai, Xiaohua and Unterthiner, Thomas and Dehghani, Mostafa and Minderer, Matthias and Heigold, Georg and Gelly, Sylvain and others},
  journal={arXiv preprint arXiv:2010.11929},
  year={2020}
}

@article{yang2025smamba,
  title={SMamba: Sparse Mamba for Event-based Object Detection},
  author={Yang, Nan and Wang, Yang and Liu, Zhanwen and Li, Meng and An, Yisheng and Zhao, Xiangmo},
  journal={arXiv preprint arXiv:2501.11971},
  year={2025}
}

@article{xiao2024spatial,
  title={Spatial-Mamba: Effective Visual State Space Models via Structure-Aware State Fusion},
  author={Xiao, Chaodong and Li, Minghan and Zhang, Zhengqiang and Meng, Deyu and Zhang, Lei},
  journal={arXiv preprint arXiv:2410.15091},
  year={2024}
}

@article{sheng2024dualmamba,
  title={Dualmamba: A lightweight spectral-spatial mamba-convolution network for hyperspectral image classification},
  author={Sheng, Jiamu and Zhou, Jingyi and Wang, Jiong and Ye, Peng and Fan, Jiayuan},
  journal={IEEE Transactions on Geoscience and Remote Sensing},
  year={2024},
  publisher={IEEE}
}

@article{guo2024mambairv2,
  title={MambaIRv2: Attentive State Space Restoration},
  author={Guo, Hang and Guo, Yong and Zha, Yaohua and Zhang, Yulun and Li, Wenbo and Dai, Tao and Xia, Shu-Tao and Li, Yawei},
  journal={arXiv preprint arXiv:2411.15269},
  year={2024}
}

@inproceedings{zamir2022restormer,
  title={Restormer: Efficient transformer for high-resolution image restoration},
  author={Zamir, Syed Waqas and Arora, Aditya and Khan, Salman and Hayat, Munawar and Khan, Fahad Shahbaz and Yang, Ming-Hsuan},
  booktitle={Proceedings of the IEEE/CVF conference on computer vision and pattern recognition},
  pages={5728--5739},
  year={2022}
}

@inproceedings{cui2023focal,
  title={Focal network for image restoration},
  author={Cui, Yuning and Ren, Wenqi and Cao, Xiaochun and Knoll, Alois},
  booktitle={Proceedings of the IEEE/CVF international conference on computer vision},
  pages={13001--13011},
  year={2023}
}

@article{kingma2014adam,
  title={Adam: A method for stochastic optimization},
  author={Kingma, Diederik P and Ba, Jimmy},
  journal={arXiv preprint arXiv:1412.6980},
  year={2014}
}

@inproceedings{gopro,
  title={Deep multi-scale convolutional neural network for dynamic scene deblurring},
  author={Nah, Seungjun and Hyun Kim, Tae and Mu Lee, Kyoung},
  booktitle={Proceedings of the IEEE conference on computer vision and pattern recognition},
  pages={3883--3891},
  year={2017}
}

@inproceedings{reside6k,
  title={FFA-Net: Feature fusion attention network for single image dehazing},
  author={Qin, Xu and Wang, Zhilin and Bai, Yuanchao and Xie, Xiaodong and Jia, Huizhu},
  booktitle={Proceedings of the AAAI conference on artificial intelligence},
  volume={34},
  number={07},
  pages={11908--11915},
  year={2020}
}

@article{lol,
  title={Deep retinex decomposition for low-light enhancement},
  author={Wei, Chen and Wang, Wenjing and Yang, Wenhan and Liu, Jiaying},
  journal={arXiv preprint arXiv:1808.04560},
  year={2018}
}

@inproceedings{rain100H,
  title={Deep joint rain detection and removal from a single image},
  author={Yang, Wenhan and Tan, Robby T and Feng, Jiashi and Liu, Jiaying and Guo, Zongming and Yan, Shuicheng},
  booktitle={Proceedings of the IEEE conference on computer vision and pattern recognition},
  pages={1357--1366},
  year={2017}
}

@inproceedings{mirnet,
  title={Learning enriched features for real image restoration and enhancement},
  author={Zamir, Syed Waqas and Arora, Aditya and Khan, Salman and Hayat, Munawar and Khan, Fahad Shahbaz and Yang, Ming-Hsuan and Shao, Ling},
  booktitle={Computer Vision--ECCV 2020: 16th European Conference, Glasgow, UK, August 23--28, 2020, Proceedings, Part XXV 16},
  pages={492--511},
  year={2020},
  organization={Springer}
}

@article{nafnet,
  title={Simple Baselines for Image Restoration},
  author={Chen, Liangyu and Chu, Xiaojie and Zhang, Xiangyu and Sun, Jian},
  journal={arXiv preprint arXiv:2204.04676},
  year={2022}
}

@inproceedings{mprnet,
    title={Multi-Stage Progressive Image Restoration},
    author={Syed Waqas Zamir and Aditya Arora and Salman Khan and Munawar Hayat
            and Fahad Shahbaz Khan and Ming-Hsuan Yang and Ling Shao},
    booktitle={CVPR},
    year={2021}
}

@article{jorder,
  title={Joint rain detection and removal from a single image with contextualized deep networks},
  author={Yang, Wenhan and Tan, Robby T and Feng, Jiashi and Guo, Zongming and Yan, Shuicheng and Liu, Jiaying},
  journal={IEEE transactions on pattern analysis and machine intelligence},
  volume={42},
  number={6},
  pages={1377--1393},
  year={2019},
  publisher={IEEE}
}

@article{maxim,
  title={MAXIM: Multi-Axis MLP for Image Processing},
  author={Tu, Zhengzhong and Talebi, Hossein and Zhang, Han and Yang, Feng and Milanfar, Peyman and Bovik, Alan and Li, Yinxiao},
  journal={CVPR},
  year={2022},
}

@article{enlightengan,
  title={Enlightengan: Deep light enhancement without paired supervision},
  author={Jiang, Yifan and Gong, Xinyu and Liu, Ding and Cheng, Yu and Fang, Chen and Shen, Xiaohui and Yang, Jianchao and Zhou, Pan and Wang, Zhangyang},
  journal={IEEE Transactions on Image Processing},
  volume={30},
  pages={2340--2349},
  year={2021},
  publisher={IEEE}
}

@InProceedings{deepdeblur,
  author = {Nah, Seungjun and Kim, Tae Hyun and Lee, Kyoung Mu},
  title = {Deep Multi-Scale Convolutional Neural Network for Dynamic Scene Deblurring},
  booktitle = {The IEEE Conference on Computer Vision and Pattern Recognition (CVPR)},
  month = {July},
  year = {2017}
}

@inproceedings{wang2023decoupling,
  title={Decoupling-and-aggregating for image exposure correction},
  author={Wang, Yang and Peng, Long and Li, Liang and Cao, Yang and Zha, Zheng-Jun},
  booktitle={Proceedings of the IEEE/CVF conference on computer vision and pattern recognition},
  pages={18115--18124},
  year={2023}
}

@article{peng2024efficient,
  title={Efficient real-world image super-resolution via adaptive directional gradient convolution},
  author={Peng, Long and Cao, Yang and Pei, Renjing and Li, Wenbo and Guo, Jiaming and Fu, Xueyang and Wang, Yang and Zha, Zheng-Jun},
  journal={arXiv preprint arXiv:2405.07023},
  year={2024}
}

@article{peng2025directing,
  title={Directing Mamba to Complex Textures: An Efficient Texture-Aware State Space Model for Image Restoration},
  author={Peng, Long and Di, Xin and Feng, Zhanfeng and Li, Wenbo and Pei, Renjing and Wang, Yang and Fu, Xueyang and Cao, Yang and Zha, Zheng-Jun},
  journal={arXiv preprint arXiv:2501.16583},
  year={2025}
}

@article{di2024qmambabsr,
  title={QMambaBSR: Burst Image Super-Resolution with Query State Space Model},
  author={Di, Xin and Peng, Long and Xia, Peizhe and Li, Wenbo and Pei, Renjing and Cao, Yang and Wang, Yang and Zha, Zheng-Jun},
  journal={arXiv preprint arXiv:2408.08665},
  year={2024}
}

@article{peng2024towards,
  title={Towards realistic data generation for real-world super-resolution},
  author={Peng, Long and Li, Wenbo and Pei, Renjing and Ren, Jingjing and Xu, Jiaqi and Wang, Yang and Cao, Yang and Zha, Zheng-Jun},
  journal={arXiv preprint arXiv:2406.07255},
  year={2024}
}

@article{peng2025pixel,
  title={Pixel to Gaussian: Ultra-Fast Continuous Super-Resolution with 2D Gaussian Modeling},
  author={Peng, Long and Wu, Anran and Li, Wenbo and Xia, Peizhe and Dai, Xueyuan and Zhang, Xinjie and Di, Xin and Sun, Haoze and Pei, Renjing and Wang, Yang and others},
  journal={arXiv preprint arXiv:2503.06617},
  year={2025}
}

@article{peng2024lightweight,
  title={Lightweight adaptive feature de-drifting for compressed image classification},
  author={Peng, Long and Cao, Yang and Sun, Yuejin and Wang, Yang},
  journal={IEEE Transactions on Multimedia},
  volume={26},
  pages={6424--6436},
  year={2024},
  publisher={IEEE}
}

\vfill
\end{document}